\documentclass[compsoc,twoside,journal]{IEEEtran}
\usepackage{algorithmicx,algorithm}
\usepackage[noend]{algpseudocode}
\usepackage{multirow}
\usepackage{array}
\usepackage{amssymb}
\usepackage{booktabs}
\usepackage[switch]{lineno}
\usepackage{color}

\ifCLASSOPTIONcompsoc
\else
  \usepackage{cite}
\fi
\ifCLASSINFOpdf
   \usepackage[pdftex]{graphicx}
\else
\fi
\usepackage[cmex10]{amsmath}
\begin{document}
%
\title{Do Inpainting Yourself: Generative\\Facial Inpainting Guided by Exemplars}
%
%
%
%

\author{Wanglong~Lu,
        Hanli~Zhao$^{*}$,
        Xianta~Jiang,
        Xiaogang~Jin,
        Yongliang~Yang,\\
        Min~Wang,
        Jiankai~Lyu,
        and~Kaijie~Shi
}

\IEEEtitleabstractindextext{%
\begin{abstract}
  We present EXE-GAN, a novel exemplar-guided facial inpainting framework using generative adversarial networks. Our approach can not only preserve the quality of the input facial image but also complete the image with exemplar-like facial attributes. We achieve this by simultaneously leveraging the global style of the input image, the stochastic style generated from the random latent code, and the exemplar style of exemplar image. We introduce a novel attribute similarity metric to encourage networks to learn the style of facial attributes from the exemplar in a self-supervised way. To guarantee the natural transition across the boundaries of inpainted regions, we introduce a novel spatial variant gradient backpropagation technique to adjust the loss gradients based on the spatial location. Extensive evaluations and practical applications on public CelebA-HQ and FFHQ datasets validate the superiority of EXE-GAN in terms of the visual quality in facial inpainting.
\end{abstract}

}

\maketitle

\IEEEdisplaynontitleabstractindextext

%
\IEEEpeerreviewmaketitle

\begin{figure*}
	\includegraphics[width=\textwidth]{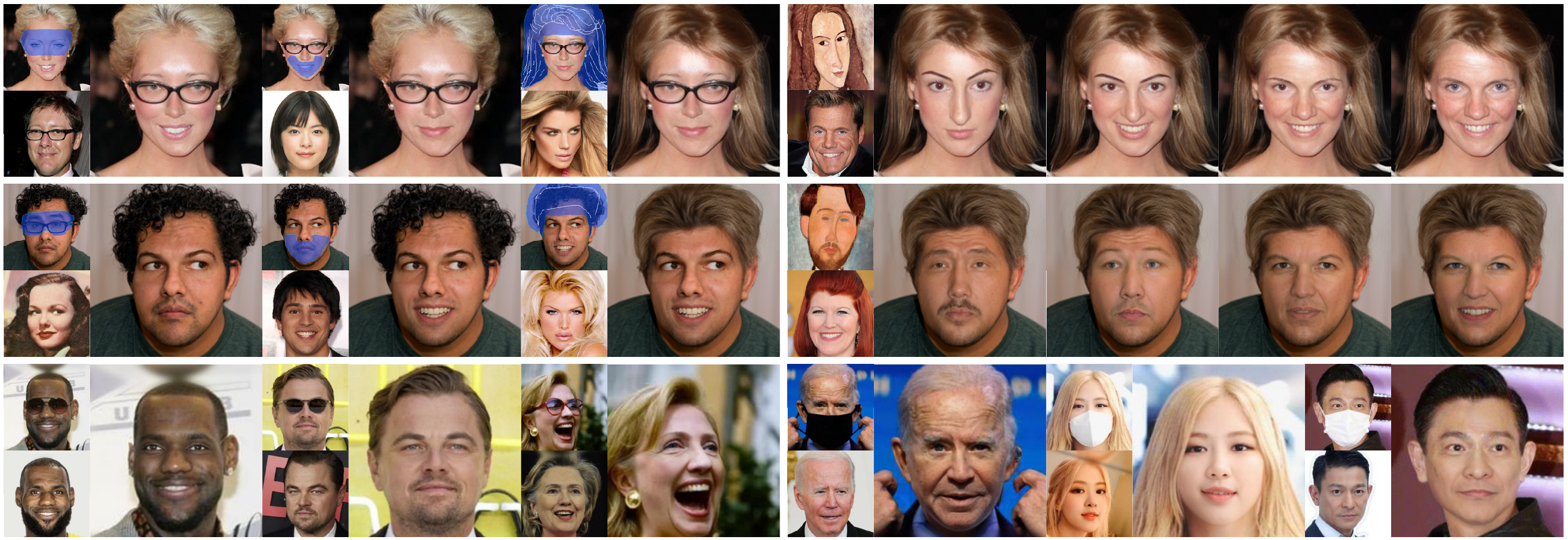}
	\caption{Facial inpainting examples using our method. Top two rows: starting with the input image (the top-left sub-image with mask), our method gradually edits the eye style (left), the mouth style (middle left), the hair style (middle), and the facial styles (right) from exemplars. Hairstyles can be edited with the insertion of basic sketches (middle). Real-world and artistic face photos can both be used to direct the inpainting of (blended) facial features in the local edited regions without affecting the visual content of the rest of the image. Bottom row: For occluded portraits with eyeglasses and masks, we perform guided facial image recovery from exemplars.}
	\label{fig:teaser}
\end{figure*}

%
%

%
%
%
%

\section{Introduction}
\label{sect:intro}  

\IEEEPARstart{F}{aces} are widely recognized as the most representative and expressive aspect of human beings. With the advancement of digital imaging and mobile computing techniques, facial photographs may now be readily collected and distributed. This increases the need for effective and fast facial image altering in a convenient manner while keeping authenticity.

In this paper, we aim to solve a new face image manipulation problem. 
The goal is to seamlessly fill in the
missing region of an input image by referring to the corresponding
content of an exemplar image. This can largely help to generate a
satisfactory face image that would favor various application scenarios,
including recovering faces occluded by face masks, sunglasses,
etc.; synthesizing faces of interest for person identification;
designing personalized hairstyles according to existing examples;
and generating face makeups for visual effects, to name just a few.

Many face image manipulation methods can achieve impressive manipulation of facial attributes based on guidance information, such as geometries~\cite{Lee2020,Chen2021}, semantics~\cite{Choi2020}, and exemplars~\cite{Li2021}. However, these methods often introduce unwanted changes to unedited regions and thus cannot guarantee visual information of known regions unchanged. 


Facial inpainting plays an important role in facial image editing for filling missing or masked regions. To achieve realistic facial inpainting guided by exemplar images, there are two main challenges: how to learn the style of facial attributes from the exemplar and how to guarantee natural transition on the mask boundary. Some works~\cite{Zhao2020,Zhao2021} attempt to generate diverse image inpainting results allowing users to select a desired one. But they cannot complete missing regions with user guidance. Many recent methods try to employ additional landmarks~\cite{Yang2020}, strokes~\cite{Jo2019}, or sketches~\cite{Yu2019,Nazeri2019} to guide the inpainting of facial structures and attributes. However, these methods tend to overfit the resulting images with these limited guidance information. As a result, these methods still require considerable professional skills in order to generate satisfactory target facial attributes, such as identity, expression, and gender.


To this end, we propose EXE-GAN, a novel  interactive  facial inpainting framework, which enables high-quality generative facial inpainting guided by exemplars. 
Our framework consists of four main components, including a mapping network, a style encoder, a multi-style generator, and a discriminator. 
Our method mixes the global style of input image, the stochastic style {generated from the random latent code}, and the exemplar style of exemplar image to generate high realistic images. We impose a perceptual similarity constraint to preserve the global visual consistency of the image. To enable the completion of exemplar-like facial attributes, we further employ facial identity and attribute constraints on the output result. To guarantee natural transition across the boundary of inpainted region, we devise a novel spatial variant gradient backpropagation method for the network training.  
We compare our method to the state-of-the-art methods to validate its advantages. Experimental results show that our method outperforms competitive methods in terms of visual quality. 
We also demonstrate several applications that could benefit from our framework, including local facial attribute transfer, guided facial style mixing, hairstyle editing, and guided facial image recovery (see Fig.~\ref{fig:teaser}). 


In summary, our paper makes the following contributions:
\begin{itemize}
	\item A novel interactive facial inpainting framework for high-quality generative inpainting of facial images with facial attributes guided by exemplars.
	
	\item A self-supervised attribute similarity metric to encourage the generative network to learn the style of facial attributes from exemplars.
	
	\item A novel spatial variant gradient backpropagation method for network training to guarantee realistic inpainting with natural transition on the boundary.
	
	\item {Several applications benefiting from the proposed facial inpainting approach, including local facial attribute transfer, guided facial style mixing, hairstyle editing, and guided facial image recovery.}
	
	
\end{itemize}

\begin{figure*}[t]
	\centering
	\includegraphics[width=\textwidth]{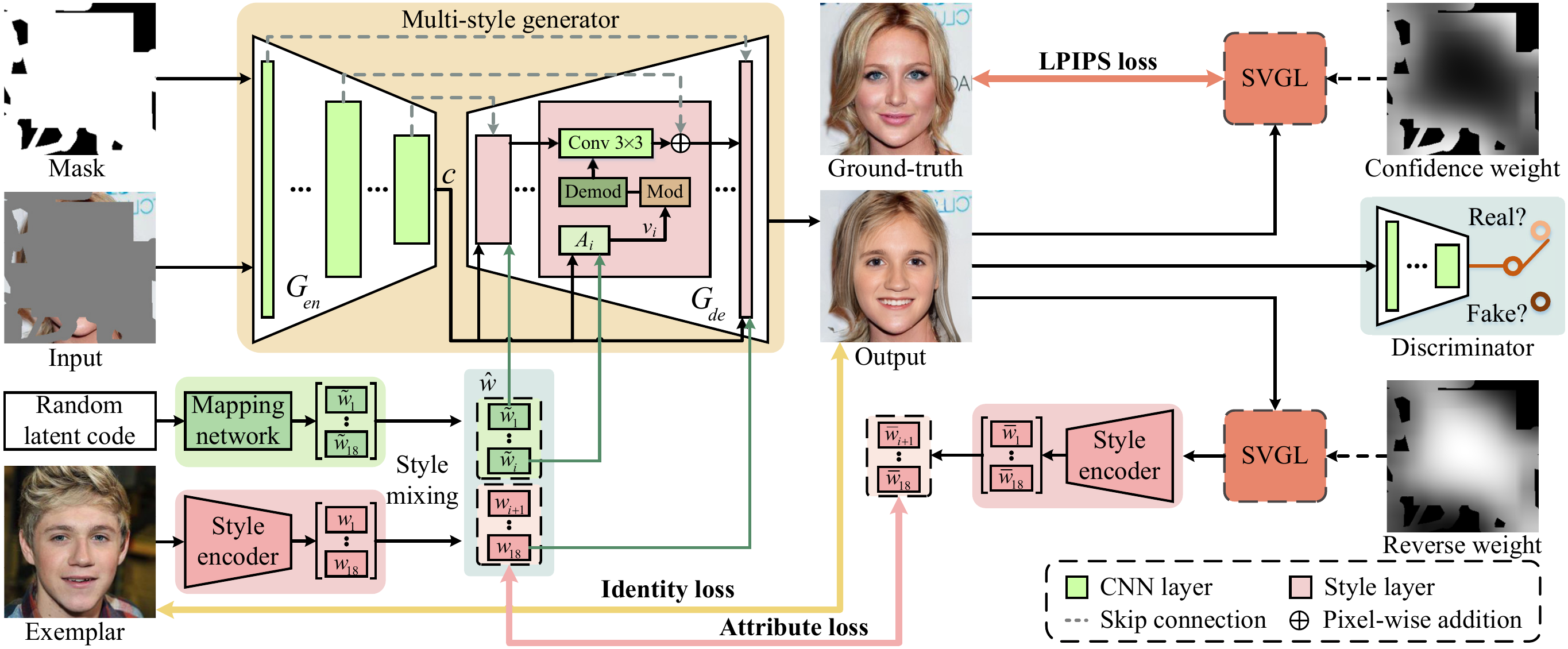}
	\caption{Overview of our EXE-GAN framework. We employ style mixing on stochastic and exemplar style codes, and modulate them with the global style code of input image into the multi-style generator for facial inpainting. The adversarial, identity, LPIPS, and attribute losses are integrated as the overall training objective. Spatial variant gradient layers (SVGL) are utilized for natural transition across the filling boundary.}\label{fig:architecture}
\end{figure*}

\section{Related work}\label{sec:prev}

\textit{Traditional image inpainting} techniques, such as {diffusion-based methods~\cite{Bertalmio2000,Levin2003} and patch-based methods~\cite{kwatra2005,Barnes2009,Zhao2018}} mainly leverage low-level features to inpaint missing regions. These methods can perform well for small and narrow missing regions, but lack semantic understanding of the image.

\textit{Deep-learning-based inpainting} methods employ deep neural networks and generative adversarial networks (GANs)~\cite{Goodfellow2014} to achieve semantic completion, such as completion with global and local discriminators~\cite{Iizuka2017}, {attention-based inpainting~\cite{Yu2018,Liu2019,Xie2019}}, inpainting via multi-column CNNs~\cite{Wang2018}, inpainting for irregular holes~\cite{Liu2018,Yu2019}, and high-resolution inpainting~\cite{Yi2020}. {Some recent works~\cite{Zheng2019,Zhao2020,Zhao2021,wan2021high,Yu2021,liu2022reduce} try to generate pluralistic inpainted images without guidance information. For instance, Co-Mod-GAN~\cite{Zhao2021} succeeds in completing large-scale missing regions and obtains diverse results by introducing inherent stochasticity. } By taking advantage of auxiliary information, such as landmarks~\cite{Yang2020}, sketches~\cite{Yu2019,Nazeri2019,Jo2019}, and geometries~\cite{Ren2019,Xiong2019}, structures and attributes of inpainted regions can be generated accordingly. Since facial attributes contain rich visual information such as color, geometry and texture, these methods tend to overfit with simple geometric information. Thus, the variety of completed solutions is quite limited. Different from these methods, our method can effectively generate realistic inpainted face images guided by an exemplar image containing rich textural and semantic information.


\textit{Facial attribute transfer} can be achieved by face manipulation methods using latent-guided codes or reference images. Semantic-level face manipulation methods can control a set of attributes, e.g., with or without glasses, mainly using a domain label to index the mapped style codes~\cite{Choi2018,Choi2020}.  However, these methods only operate on a set of pre-defined attributes and leave users little freedom for face manipulation. Geometry-based face manipulation methods implement exemplar-based facial transfer based on semantic geometry~\cite{Lee2020,Chen2021}.  Since information loss occurs in the projection and reconstruction between real-captured photos and corresponding representations, these methods tend to change fine details in the background. Exemplar-based face manipulation methods transfer facial attributes from exemplars at the instance level~\cite{Xiao2018,Guo2019,Li2020_face,Chen2020_simSwap,Li2021,Liu_2021_CVPR}. {In particular, SimSwap~\cite{Chen2020_simSwap} extends the identity-specific face swapping to arbitrary face swapping. }
Nevertheless, these methods have a common limitation as users cannot flexibly select facial regions for the local transfer of facial attributes. {Kim et al.~\cite{kim2021stylemapgan} proposed an intermediate representation with spatial dimensions to perform local editing, but their method may fail when the poses in the original and reference images differ.} In contrast, our method transfers local facial attributes from exemplars interactively {to produce realistic inpainted face images with natural attribute transitions while remaining unaltered regions.} 

Our work is also closely related to \textit{image embedding} which enables image synthesis {from} latent space. StyleGANs~\cite{Karras2019,Karras2020} enable direct scale-specific control of image synthesis with disentangled intermediate latent style space and can produce plausible results for unconditional face image synthesis. {Optimization-based embedding~\cite{Abdal2019,Abdal_2020_cvpr} and encoder-based embedding~\cite{Zhu2020,Richardson2021,Tov2021,Wu2021,Wu_2022_CVPR} methods perform image manipulation by inverting an image to the latent space~\cite{Zhu2016}.} {Recently, EditingInStyle~\cite{Collins2020} and StyleFusion~\cite{Kafri2022} show the impressive performance of image editing by semantically manipulating the style latent space. Although image embedding has the strong capability in presenting image styles, these methods may change unedited regions because of information losses in GAN inversion. For instance, Richardson et al.~ \cite{Richardson2021} presented results of inpainting using a pixel2style2pixel framework but failed to preserve visual contents of unmasked parts. 
} In this work, we propose a novel facial inpainting framework by taking advantage of style latent codes {while keeping the unmasked region.}



\section{Method}\label{sec:algo}

\subsection{Overview}

The overall structure of our EXE-GAN framework is shown in Fig.~\ref{fig:architecture}. Given a ground-truth face image $I_{gt} \in \mathbb{R}^{h \times w \times 3}$, an exemplar image $I_{exe} \in \mathbb{R}^{h \times w \times 3}$, and a binary mask $M \in \mathbb{R}^{h \times w \times 1}$ (with value 1 for unknown and 0 for known pixels), the input image $I_{in} \in \mathbb{R}^{h \times w \times 3}$ is obtained by $I_{in} = I_{gt} \odot (1-M)$, where $\odot$ denotes the Hadamard product. The goal of our EXE-GAN framework is to automatically generate a realistic face image $I_{out}$, where the inpainting of the masked regions in $I_{in}$ is guided by the facial attributes of $I_{exe}$ while the known regions remain unchanged. The proposed EXE-GAN consists of four main components, including a mapping network, a style encoder, a multi-style generator, and a discriminator.

\subsubsection{Mapping network} A multi-layer fully-connected neural network $f$~\cite{Karras2020,Zhao2021} linearly maps {a random latent code  $z \in \mathbb{R}^{512 \times 1}$} to a stochastic style code $\widetilde{w} = \left\lbrace \widetilde{w}_{i} \in \mathbb{R}^{512 \times 1} | i \in T \right\rbrace \in \widetilde{W}+$, where $\widetilde{W}+$ denotes the extended stochastic style latent space and $T = \lbrace1, 2,..., 18\rbrace$ denotes the index set. Let $\theta_{f}$ be the learnable network parameters in $f$, we have $\widetilde{w} = f(z;\theta_{f})$.

\subsubsection{Style encoder} A pre-trained pixel2style2pixel style encoder $E$~\cite{Richardson2021,Tov2021} directly maps an image to a disentangled style latent space $W+$. Given the pre-trained network parameters $\hat{\theta}_{e}$, the style encoder extracts the exemplar style code $w = \left\lbrace w_{i} \in \mathbb{R}^{512 \times 1} | i \in T \right\rbrace = E(I_{exe};\hat{\theta}_{e})$ and the inpainted style code $\overline{w} = \left\lbrace \overline{w}_{i} \in \mathbb{R}^{512 \times 1} | i \in T \right\rbrace = E(I_{out};\hat{\theta}_{e})$. Therefore, $w \in W+$ and $\overline{w} \in W+$.

\subsubsection{Multi-style generator} A generative network $G$ that leverages multiple representations (i.e., $I_{in}$, $M$, and $\hat{w}$) to generate an intermediate result $I_{pred} \in \mathbb{R}^{h \times w \times 3}$, where $\hat{w} = \left\lbrace \hat{w}_{i} \in \mathbb{R}^{512 \times 1} | i \in T \right\rbrace$ is the mixed style code of ${w}$ and $\widetilde{w}$. Let $\theta_{g}$ be the learnable network parameters of $G$, we have $I_{pred} = G(I_{in}, M,\hat{w};\theta_{g})$. The multi-style generator can be further divided into an encoder $G_{en}$ and a decoder $G_{de}$, i. e., $G= {\{G_{en},G_{de}\}}$.

\subsubsection{Discriminator} A discriminative network $D$ \cite{Karras2020,Zhao2021} learns to judge whether an image is a real or fake image. Let $\theta_{d}$ be the learnable network parameters of $D$, the discriminative network maps the inpainted image $I_{out}$ to a scalar $D(I_{out};\theta_{d}) \in \mathbb{R}^{1 \times 1}$.

\subsection{Multi-style modulation}\label{sec:multi_style}

To leverage the global style of the input image, the stochastic style {generated from the random latent code}, and the exemplar style of exemplar image to perform generative facial inpainting, we build upon the generator of Co-mod-GAN~\cite{Zhao2021} and
extend it to our multi-style generator $G$. This is done by incorporating also the exemplar style, and mixing it with other styles based on carefully designed style modulation. The proposed multi-style generator can not only preserve the global visual consistency of the input image, but also embed exemplar facial attributes to the local facial inpainting. In addition, it has the good property of inherent stochasticity with the stochastic style latent code.

First of all, the mixed style code $\hat{w}$ is obtained by style mixing~\cite{Karras2019,Karras2020} of the stochastic and exemplar styles. Specifically, each layer of $\hat{w}$ is defined as:
\begin{equation}\label{equ:mixing}
\begin{aligned}
\hat{w}_{i} = \begin{cases} w_{i}, & \text {if } \phi_{i} = 1,
\\\widetilde{w}_{i}, & \text {otherwise,} \end{cases}
\end{aligned}
\end{equation}
where $i \in T = \lbrace1, 2,..., 18\rbrace$ and $\phi \in \mathbb{R}^{18 \times 1}$ is a binary vector to indicate which style is modulated for each layer. {As demonstrated in StyleGAN~\cite{Karras2019}, coarse-resolution layers correspond to high-level facial attributes and fine-resolution layers could change small-scale features. We empirically set $\phi = [0,0,0,0,1,1,...,1]$ by balancing the stochastic and exemplar styles in this paper.}

Secondly, the encoder $G_{en}$ takes $I_{in}$ and $M$ as input, and outputs a global style code $c \in \mathbb{R}^{2 \times 512 \times 1}$ as well as the corresponding multi-resolution feature maps.

Then, as illustrated in Fig.~\ref{fig:architecture}, the global style code $c$ and the mixed style code $\hat{w}$ are transformed to multi-style vectors $v$ for subsequent modulation within the style layers of the decoder $G_{de}$. For each $i$-th style layer, the transformation is defined as~\cite{Karras2020}:
\begin{equation}\label{equ:affine}
\begin{aligned}
v_{i}={A}_{i}([c, \hat{w}_{i} ]),
\end{aligned}
\end{equation}
where $[\cdot]$ refers to the concatenation operator, ${A}_{i}$ is a learned affine transformation within the $i$-th style layer, and $v_{i}$ is a linearly learned style representation conditioned on the input style representations.

Next, the decoder $G_{de}$ utilizes the multi-style vectors $v$ and the multi-resolution feature maps output by $G_{en}$ to generate the intermediate inpainting $I_{pred}$. The decoder contains two style layers in each resolution. In each $i$-th style layer, the multi-style vector $v_{i}$ is then used for weight modulation and demodulation~\cite{Karras2020,Zhao2021}. As shown in Fig.~\ref{fig:architecture}, skip connections are used for collecting the multi-resolution feature maps in the decoder $G_{de}$.

Finally, the inpainted image $I_{out}$ is generated as follows:
\begin{equation}\label{equ:output}
\begin{aligned}
I_{out} = I_{in} \odot (1 - M) + I_{pred} \odot M.
\end{aligned}
\end{equation}

\subsection{Training objectives}\label{sec:co-modulated}

Our framework is trained to optimize the learnable network parameters $\theta_{g}$, $\theta_{f}$, and $\theta_{d}$ using the following objectives.

\subsubsection{Adversarial loss}
We use the adversarial non-saturating logistic loss~\cite{Goodfellow2014} with $R_{1}$ regularization~\cite{Mescheder2018}. Specifically, the adversarial objective is defined as:
\begin{equation}\label{equ:loss_adv}
\begin{aligned}
\mathcal{L}_{adv}(I_{out},I_{gt}) = \mathbb{E}_{I_{out}}[\log(1-D(I_{out})] \\+\mathbb{E}_{I_{gt}}[\log(D(I_{gt}))] - \frac{\gamma}{2} \mathbb{E}_{I_{gt}}[\|  \nabla_{I_{gt}}& D(I_{gt})\|_{2}^{2}],\\
\end{aligned}
\end{equation}
where $\gamma$ is used to balance the $R_1$ regularization term. We empirically set $\gamma = 10$. The generative network $G$ learns to generate a visually realistic image $I_{out}$ while the discriminative network $D$ tries to distinguish between the ground-truth $I_{gt}$ and the generated image $I_{out}$. $G$ and $D$ are trained in an alternating manner.

\subsubsection{Identity loss}
We constrain identity similarity between the output image $I_{out}$ and the exemplar image $I_{exe}$ in the embedding space. The identity loss is formulated as follows: 
\begin{equation}\label{equ:loss_id}
\begin{aligned}
\mathcal{L}_{id}(I_{out},I_{exe})= 1- \cos\left(R(I_{out}),R(I_{exe})\right),
\end{aligned}
\end{equation}
where $R(\cdot)$ is a pre-trained ArcFace network~\cite{Deng2019} for face recognition.

\subsubsection{LPIPS loss}\label{sec:LPIPS}
We employ the the Learned Perceptual Image Patch Similarity (LPIPS) loss~\cite{Zhang2018} to constrain the perceptual similarity between the output image $I_{out}$ and the ground-truth $I_{gt}$:
\begin{equation}\label{equ:loss_lpips}
\mathcal{L}_{lpips}(I_{out},I_{gt})= \begin{cases}\| F(I_{out})-F(I_{gt})\|_{2} & \text { if } I_{gt}=I_{exe} \\ 0 & \text { otherwise }\end{cases},
\end{equation}
where $F(\cdot)$ is the pre-trained perceptual feature extractor and we adopt VGG~\cite{Simonyan2014} in our work. Note that $\mathcal{L}_{lpips}$  is applied only when $I_{gt}$ and $I_{exe}$ are sampled from the same image (see Section~\ref{sec:exp_settings} for the detailed settings). 

\subsubsection{Attribute loss}\label{sec:attribute_loss}

In order to learn the style of facial attributes from the exemplar image, we introduce a novel self-supervised attribute similarity metric to measure the consistency between facial attributes of the inpainted result $I_{out}$ and the exemplar $I_{exe}$ in the style latent space:
\begin{equation}\label{equ:loss_attri}
\begin{aligned}
\mathcal{L}_{attr}(I_{out},I_{exe}) = \frac{1}{\|\phi\|_{0}}\sum_{i \in T} \phi_{i} \cdot \| \overline{w}_{i}  - \hat{w}_{i} \|_{2},
\end{aligned}
\end{equation}
where the L0 norm $\| \cdot \|_{0}$ indicates the number of non-zeros.

\subsubsection{Total objective}\label{sec:total_losses}

After defining loss functions above, the total training objective can be expressed as:
\begin{equation}\label{equ:loss_total}
\begin{aligned}
{O}(\theta_{g},\theta_{f},\theta_{d}, \hat{\theta}_{e}) = \mathcal{L}_{adv}(I_{out},I_{gt}) + \lambda_{id}\mathcal{L}_{id}(I_{out},I_{exe}) \\ + \lambda_{lpips}\mathcal{L}_{lpips}(I_{out},I_{gt}) +\lambda_{attr}\mathcal{L}_{attr}(I_{out},I_{exe}),
\end{aligned}
\end{equation}
where $\lambda_{id}$, $\lambda_{lpips}$, and $\lambda_{attr}$ are weights of corresponding losses, respectively. We empirically set $\lambda_{id}=0.1$, $\lambda_{lpips}=0.5$, and $\lambda_{attr}=0.1$ in this work. During training, we can obtain the optimized parameters ${\theta}_{g}$, ${\theta}_{f}$, and ${\theta}_{d}$ via the minimax game iteratively:
\begin{equation}\label{equ:loss_minimax}
\begin{aligned}
({\theta}_{g}, {\theta}_{f})&=\arg \min _{\theta_{g}, \theta_{f}} {O}(\theta_{g}, \theta_{f}, \theta_{d},\hat{\theta}_{e}),\\
({\theta}_{d})&=\arg \max _{\theta_{d}} {O}(\theta_{g}, \theta_{f}, \theta_{d},\hat{\theta}_{e}).
\end{aligned}
\end{equation}

\subsection{Spatial variant gradient backpropagation}\label{sec:gradient_variant}

It is expected that the inpainted facial attributes close to the filling center are more similar to those of the exemplar image. Moreover, the inpainted values close to the boundary should be perceptually more similar to those of input image and the visual contents should be naturally transited on the boundary. Therefore, in order to generate naturally looking inpainting, we further exert constraint based on spatial location.

From Eqs.~\ref{equ:loss_lpips} and ~\ref{equ:loss_attri}, we can find that the LPIPS loss and attribute loss are defined over the entire inpainted image. GMCNN~\cite{Wang2018} applies the spatial constraint to the pixel-wise reconstruction loss. However, we cannot directly impose GMCNN's spatial constraint on our loss functions. The reason is that our losses are defined in embedding space and dimensions of embedding features do not match those of the spatial space. In our work, a novel spatial variant gradient layer (SVGL) is designed to impose the spatial constraint on loss gradients in backpropagation.

\begin{figure}[t]
	\centering
	\includegraphics[width=0.485\textwidth]{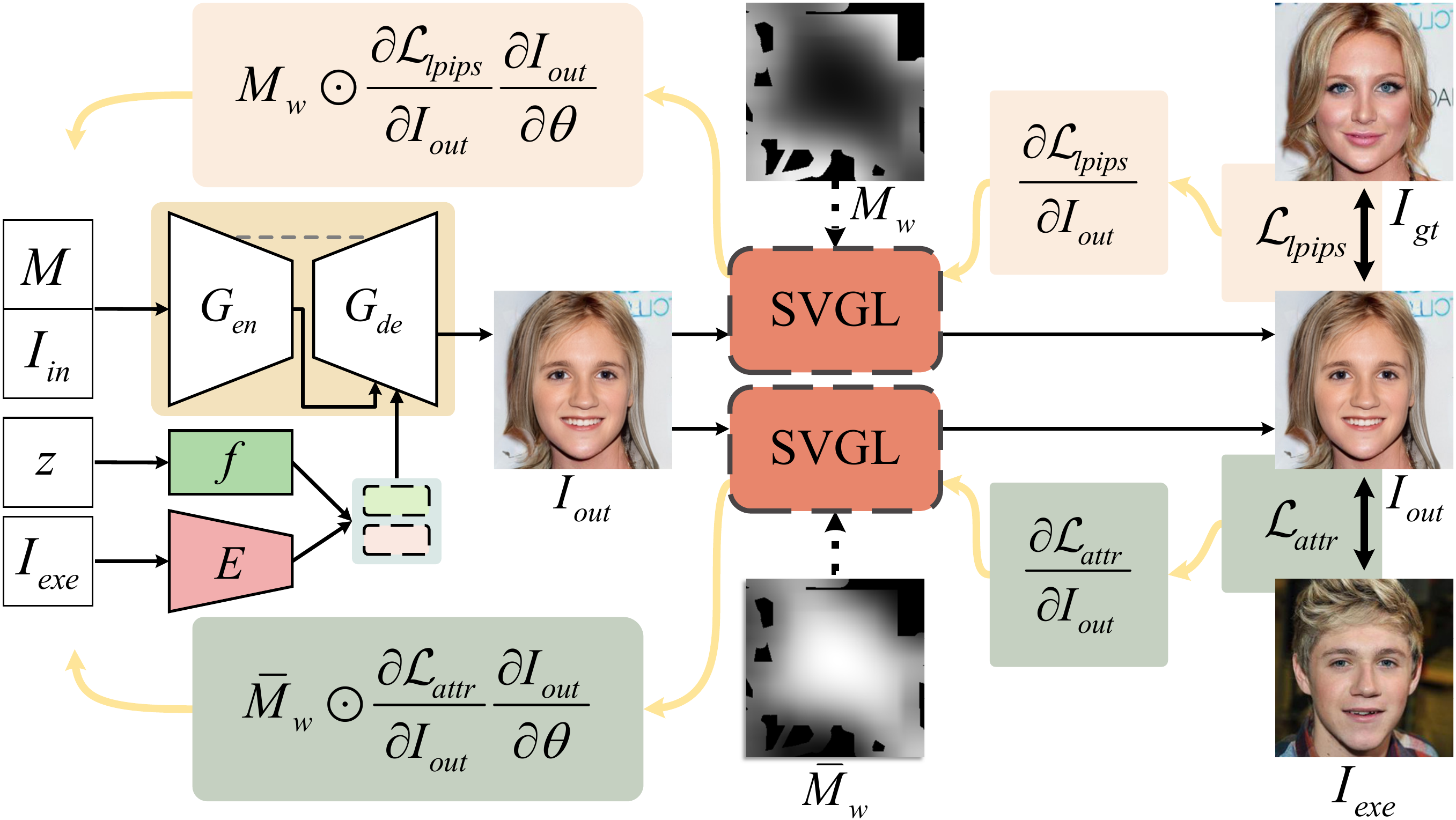}
	\caption{Illustration of the SVGL on LPIPS and attribute losses. In forward-propagation, SVGL does not change any information for $I_{out}$. In backpropagation, gradients are re-weighted based on the spatial variant $M_{w}$ and $\overline{M}_{w}$, respectively.  }\label{fig:SVGL_layer}
\end{figure}

As illustrated in Fig.~\ref{fig:SVGL_layer}, SVGL has no parameter but relies on a spatial weight mask. During forward-propagation, SVGL acts as an identity transform, which does not change any information from the input. During backpropagation, it collects gradients from subsequent layers, re-weights the gradients based on the spatial weight mask, and passes the re-weighted gradients to the preceding layers.
\begin{figure*}[t]
	\centering
	\includegraphics[width=\textwidth]{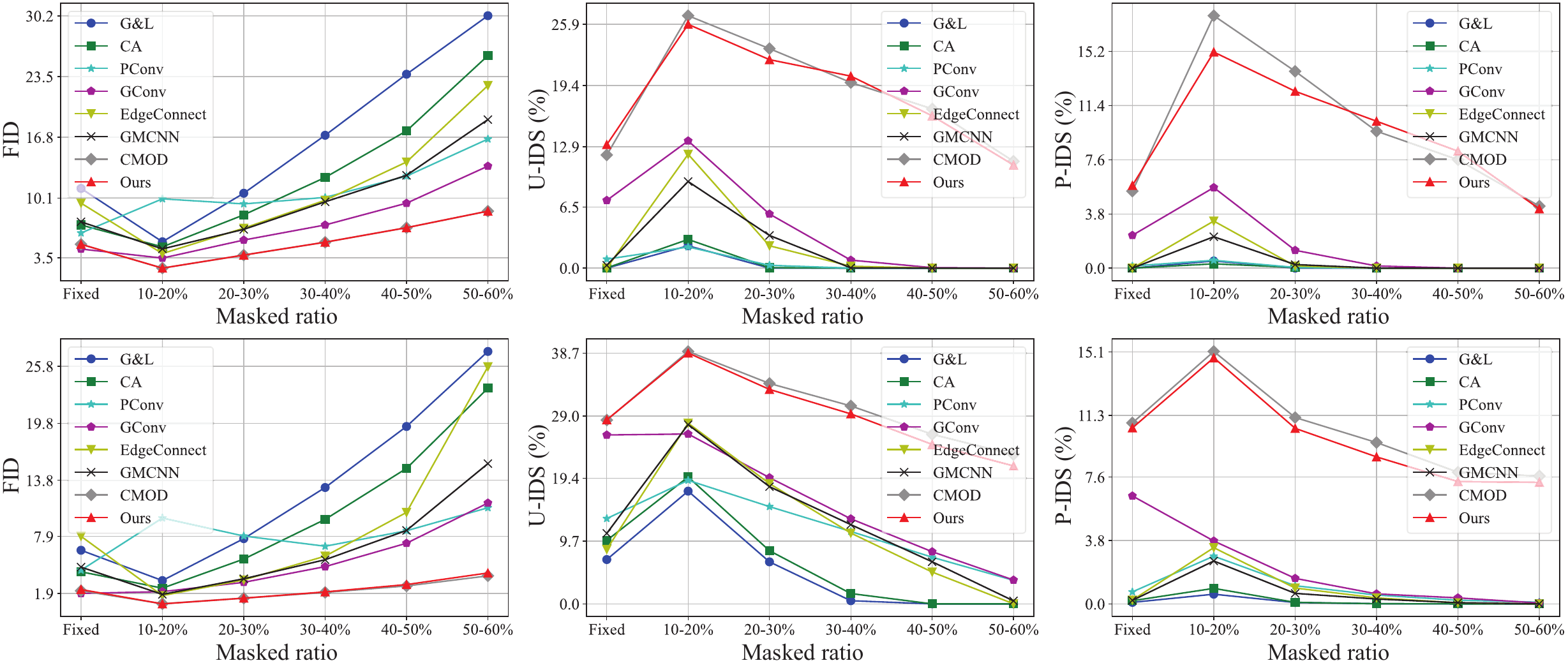}
	\caption{Quantitative comparisons between our method and the state-of-the-art free-form inpainting methods on the CelebA-HQ (top row) and FFHQ (bottom row) datasets, respectively.}\label{fig:fig_free_form_table_celeba}
\end{figure*}

Mathematically, given an input feature $x$ and a spatial weight mask $M_x$, we can treat SVGL as a ``pseudo-function'' $P(x,M_x)$. The forward-propagation and backpropagation behaviors of SVGL are defined below:
\begin{equation}\label{equ:svgl}
\begin{aligned}
P(x,{M}_{x}) &= x,\\
\frac{ \partial P( x,{M}_{x}) }{\partial x} &= {M}_{x} \odot \mathbf{I},
\end{aligned}
\end{equation}
where $\mathbf{I}$ represents an identity matrix.

Then, we apply SVGL to the spatial variant LPIPS and attribute losses. Specifically, we equipped the network with a SVGL $P(\cdot, M_w)$ for the spatial variant LPIPS loss and a SVGL $P(\cdot, \overline{M}_w)$ for the spatial variant attribute loss, respectively, where the confidence weight mask $M_{w} \in \mathbb{R}^{h \times w \times 1}$ is obtained with Gaussian smoothing on the masked region of $M$ and the reverse weight mask $\overline{M}_{w}  = (1 - M_{w}) \odot M \in \mathbb{R}^{h \times w \times 1}$. As shown in Fig.~\ref{fig:SVGL_layer}, both SVGLs are added right after the layer of generating $I_{out}$. Our SVGL is general and can be used to apply spatial constraints to any loss functions with spatial variant backpropagation. 
{Note that the values of non-masked regions are zeros and the weights of the pre-trained style encoder are frozen during training.}
With the spatial variant gradient layers, the training objective is computed with Eq.~\ref{equ:loss_total} during forward-propagation while its gradients are computed in a spatial variant manner. 

\begin{figure*}[t]
\centering
\includegraphics[width=\textwidth]{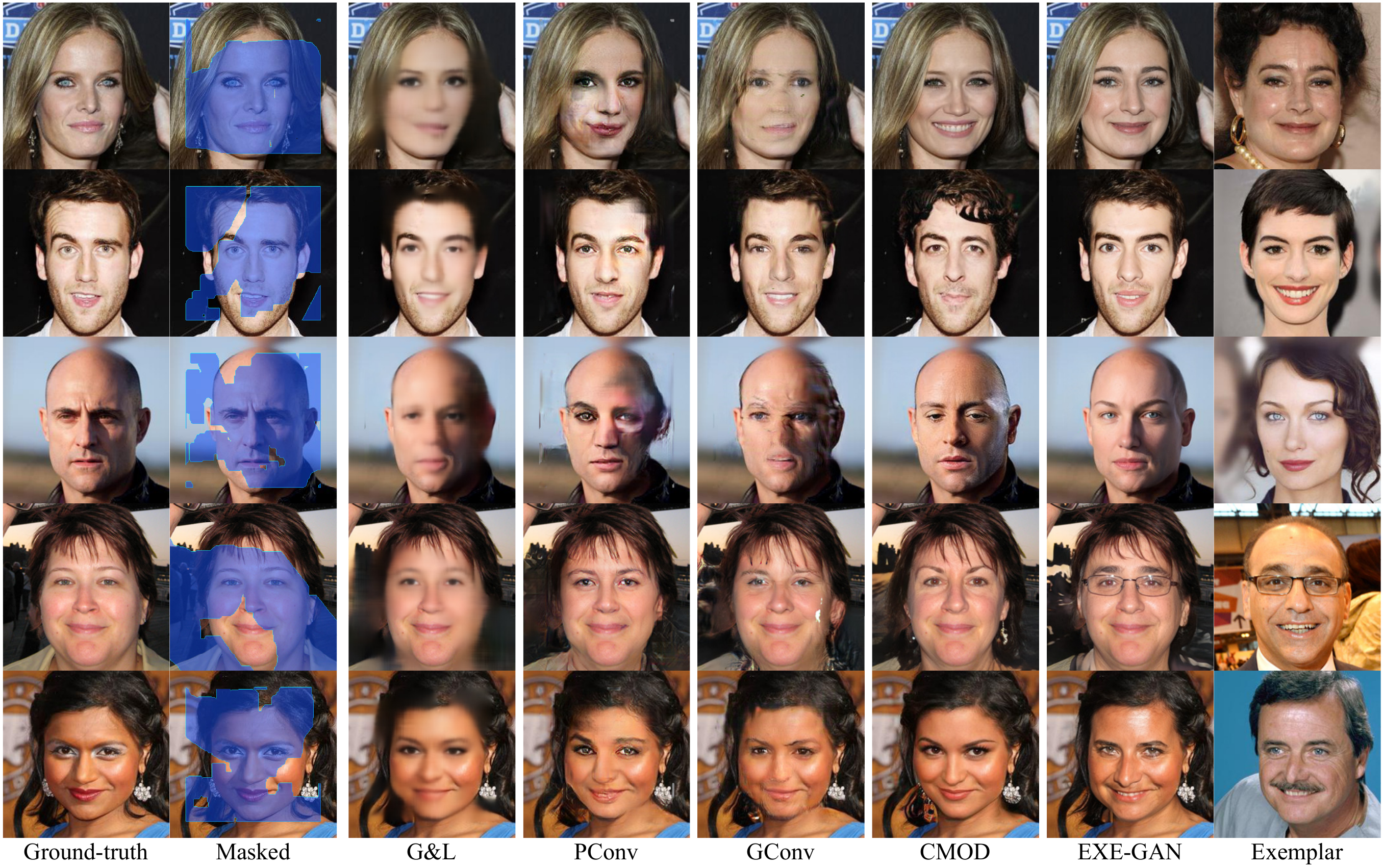}
\caption{Qualitative comparison between our method and the state-of-the-art free-form inpainting methods. }\label{fig:fig_free_form_inpainting_sub}  
\end{figure*}

\begin{algorithm}[t]
	\caption{Training procedure of EXE-GAN}
	\label{procedure_framework}
	\begin{algorithmic}[1]
		\While {$f$, $G$, and $D$ have not converged}
		\State Sample batch images $\mathcal{I}_{gt}$ from training data
		\State Sample random latent vectors $\mathcal{Z}$
		\State Sample a random number $r \in [0,1]$
		\If {$r > $ threshold $\tau$}
		\State Sample batch exemplars $\mathcal{I}_{exe}$ from training data
		\Else
		\State Set batch exemplars from ground-truth $\mathcal{I}_{exe} \leftarrow \mathcal{I}_{gt}$
		\EndIf
		\State Create random masks $\mathcal{M}$ for $\mathcal{I}_{in}$
		\State Get confidence weight masks $\mathcal{M}_{w}$ for $\mathcal{L}_{lpips}$
		\State Get reverse weight masks $\overline{\mathcal{M}}_{w}$ for $\mathcal{L}_{attr}$
		\State Get inputs $\mathcal{I}_{in} \leftarrow \mathcal{I}_{gt} \odot (1 - \mathcal{M}) $
		\State Get $\hat{w} \leftarrow mixing(E(\mathcal{I}_{exe}),f(\mathcal{Z}))$
		\State Get $\mathcal{I}_{pred} \leftarrow G\left(\mathcal{I}_{in}, \mathcal{M}, \hat{w} \right)$
		\State Get outputs $\mathcal{I}_{out} \leftarrow  \mathcal{I}_{in} \odot (1 - \mathcal{M})+ \mathcal{I}_{pred} \odot \mathcal{M}$
		\State Update $f$ and $G$ with  $\mathcal{L}_{adv}$, $\mathcal{L}_{id}$, $\mathcal{L}_{lpips}$, and $\mathcal{L}_{attr}$
		\State Update $D$ with  $\mathcal{L}_{adv}$
		\EndWhile
	\end{algorithmic}
\end{algorithm}

\section{Experiments}\label{sec:evaluation}

\subsection{Settings}
\label{sec:exp_settings}

\subsubsection{Datasets}
Experiments were conducted on two publicly available face image datasets CelebA-HQ~\cite{Karras2018} and FFHQ~\cite{Karras2019}. For CelebA-HQ~\cite{Karras2018}, we randomly selected 28,000 images for training and remained 2,000 images for testing. For FFHQ~\cite{Karras2019}, we randomly selected 60,000 images for training and the rest 10,000 images for testing. Each image was resized to $256 \times 256$. 

\subsubsection{{Evaluation metrics}}
{The performance was quantitatively evaluated using the Fréchet inception distance (FID)~\cite{Heusel2017} and the paired/unpaired inception discriminative score (P-IDS/U-IDS)~\cite{Zhao2021}. FID has been proven to correlate well with human perception for the visual quality of generated images. P-IDS and U-IDS are robust assessment measures for the perceptual fidelity of generative models.}

\subsubsection{Implementation details}

Algorithm~\ref{procedure_framework} lists the pseudo-code for our EXE-GAN framework's training procedure. The threshold $\tau \in [0,1]$ was used to control the probability that the sampled ground-truth image and exemplar image were the same. We set threshold $\tau=0.1$ in this paper.
Our framework was implemented using Python and PyTorch. Following the settings of StyleGANv2~\cite{Karras2020} and Co-mod-GAN~\cite{Zhao2021}, we employed the Adam optimizer with the first momentum coefficient of $0.5$, the second momentum coefficient of $0.99$, and the learning rate of $0.002$. Mixing regularization~\cite{Karras2019} with a probability of $0.5$ was employed to generate stochastic style codes during training. The free-form mask sampling strategy was adopted for training by simulating random brush strokes and rectangles. The brush strokes were generated using the algorithm presented in GConv~\cite{Yu2019} with maxVertex of $20$, maxLength of $100$, maxBrushWidth of $24$, and maxAngle of $360$. Multiple up-to-half-size rectangles and up-to-quarter-size rectangles were generated randomly. The numbers of up-to-half-size rectangles and up-to-quarter-size rectangles were uniformly sampled within $[0, 5]$ and $[0, 10]$, respectively.
We trained the networks for 800,000 iterations with batch size of 8. All experiments were conducted on the NVIDIA Tesla V100 GPU. The training time was around three weeks.

\begin{figure*}[t]
	\centering
	\includegraphics[width=\textwidth]{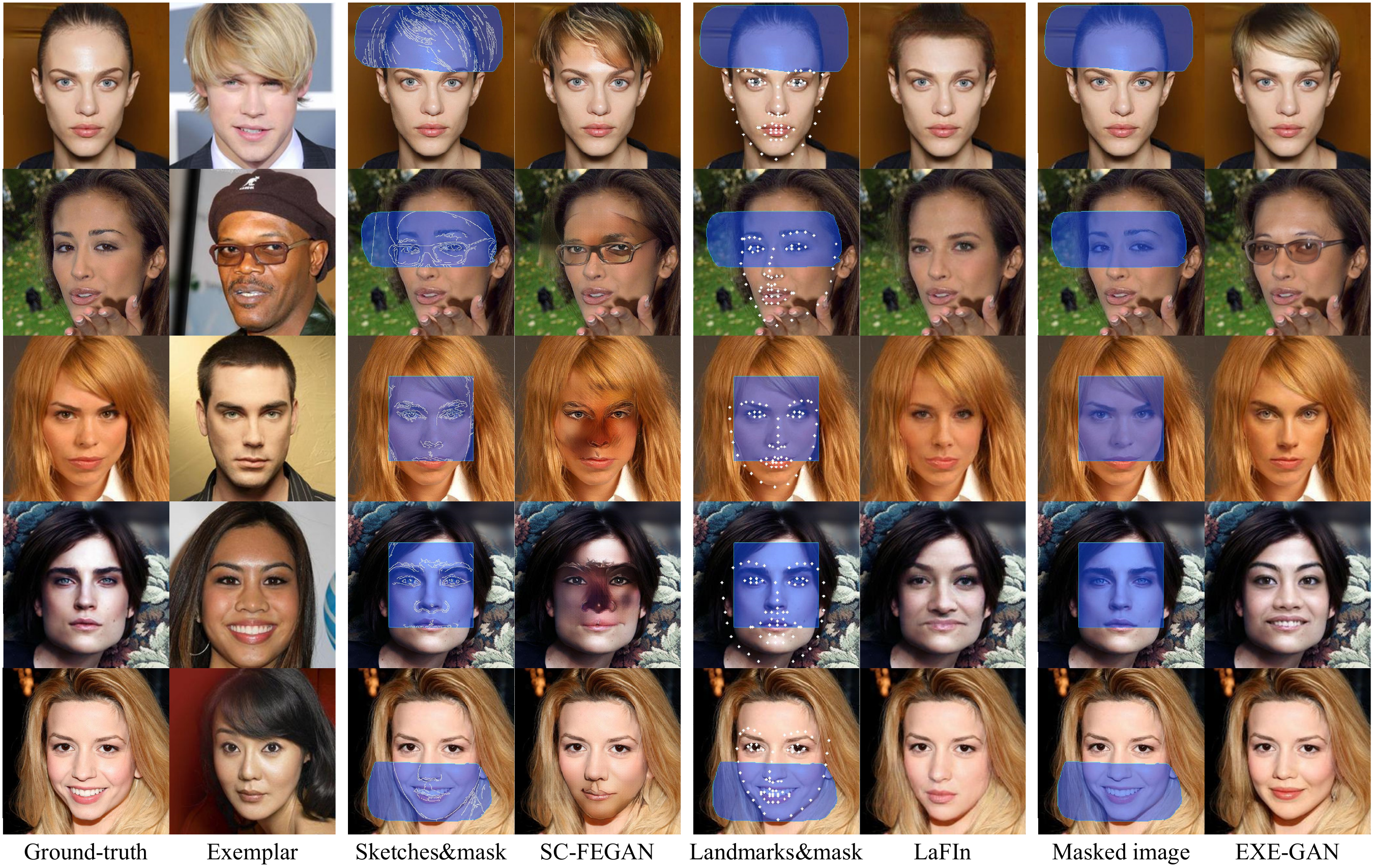}
	\caption{Qualitative comparison between our method and the state-of-the-art guidance-based facial inpainting methods.}\label{fig:attribute_inpainting_sub}  
\end{figure*}

\subsection{Comparison to free-form inpainting}\label{sec:comparison_free_form}

We compared EXE-GAN on CelebA-HQ and FFHQ datasets to the state-of-the-art free-form inpainting methods, including Contextual Attention (CA)~\cite{Yu2018}, Partial Convolutions (PConv)~\cite{Liu2018}, Globally \& Locally (G\&L)~\cite{Iizuka2017}, Gated Convolution (GConv)~\cite{Yu2019}, EdgeConnect~\cite{Nazeri2019}, GMCNN~\cite{Wang2018}, and Co-mod-GAN (CMOD)~\cite{Zhao2021}.

\subsubsection{Experiment settings}
We used the publicly available MMEditing framework~\cite{mmediting2022} for Contextual Attention (CA)~\cite{Yu2018}, Partial Convolutions (PConv)~\cite{Liu2018}, Globally \& Locally (G\&L)~\cite{Iizuka2017}, and Gated Convolution (GConv)~\cite{Yu2019}. MMEditing is an open-source image and video editing toolbox based on PyTorch. We used the official codes for EdgeConnect~\cite{Nazeri2019} and GMCNN~\cite{Wang2018}. For Co-mod-GAN (CMOD)~\cite{Zhao2021}, we used the official TensorFlow-based code to implement a PyTorch-based version. All the compared free-form inpainting models were trained using CelebA-HQ and FFHQ datasets, respectively. {For fair comparison, we used the same training/testing splits for all experiments. We randomly took exemplar images during training while using reverse indices to the input image batch during testing.}

\subsubsection{Quantitative comparison}
Fig.~\ref{fig:fig_free_form_table_celeba} shows the quantitative performance comparisons between our method and the state-of-the-art free-form methods on CelebA-HQ and FFHQ datasets, respectively. Various irregular masks with different masked ratios as well as a fixed rectangle center mask with the size of $128 \times 128 $ were employed to simulate various situations for facial image inpainting. Quantitative results show that our method performs better than most of the compared free-form inpainting methods, even though the inpainting of our method was guided by exemplars, which was considered challenging to keep image quality~\cite{Zhao2021}.  Moreover, our EXE-GAN can achieve comparable performance to CMOD~\cite{Zhao2021} with various kinds of masks in terms of FID~\cite{Heusel2017}, U-IDS and P-IDS~\cite{Zhao2021}.

\subsubsection{Qualitative comparison}
As shown in Fig.~\ref{fig:fig_free_form_inpainting_sub}, although all the methods can compatibly fill in the missing regions, G\&L~\cite{Iizuka2017} tends to produce blurry inpainting while PConv~\cite{Liu2018} and GConv~\cite{Yu2019} fail to inpaint large-scale missing regions. Our method and CMOD~\cite{Zhao2021} can generate competitive inpainted results. However, the inpainting of facial attributes cannot be controlled with CMOD~\cite{Zhao2021}. In contrast, with the help of exemplar facial attributes, the inpainting of facial attributes with our EXE-GAN can be controlled easily.


\begin{figure}[t]
	\centering
	\includegraphics[width=0.4\textwidth]{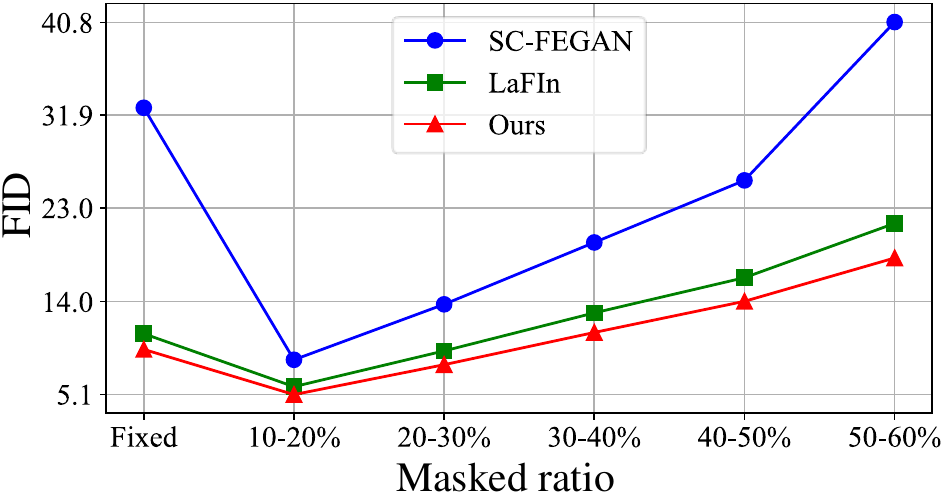}
	\caption{Quantitative comparison between our method and the state-of-the-art guidance-based facial inpainting methods on the Celeba-HQ dataest.}\label{fig:facial_inpainting}  
\end{figure}

\subsection{Comparison to guidance-based inpainting}\label{sec:comparison_guidance} 
We compared our method on the Celeba-HQ dataset to the state-of-the-art guidance-based facial inpainting methods, including sketch-and-color-based facial inpainting SC-FEGAN~\cite{Jo2019} and landmark-based face inpainting LaFIn~\cite{Yang2020}. For the compared methods, we generated corresponding guided information from exemplar images and alternately took one facial image as the exemplar and the other one as the masked image for facial attribute inpainting.

\subsubsection{Experiment settings}
The officially released pre-trained SC-FEGAN~\cite{Jo2019} and LaFIn~\cite{Yang2020} models were used in this experiment. SC-FEGAN~\cite{Jo2019} uses sketches and color as the guidance to generate missing pixels. Therefore, we leveraged the Canny edge detector to automatically generate sketches from the exemplar image. To avoid inconsistency of color in the inpainted pixels, we did not introduce color information of the exemplar into missing regions. LaFIn~\cite{Yang2020} relies on landmarks to fill missing regions. Therefore, we utilized the face alignment network FAN~\cite{Bulat2017} to generate landmarks for the exemplar image. To avoid the misalignment between the guidance information (i.e., sketches and landmarks) and unmasked regions in the masked image, we first extracted the angles of roll, pitch, and yaw from the CelebAMask-HQ dataset, then selected 550 pairs with similar poses from the testing set. For each pair, we alternately took one facial image as the exemplar and the other one as the masked image to perform facial attribute inpainting. As a consequence, we obtained 1,100 inpainted images for each comparison method.

\begin{figure*}[t]
	\centering
	\includegraphics[width=\textwidth]{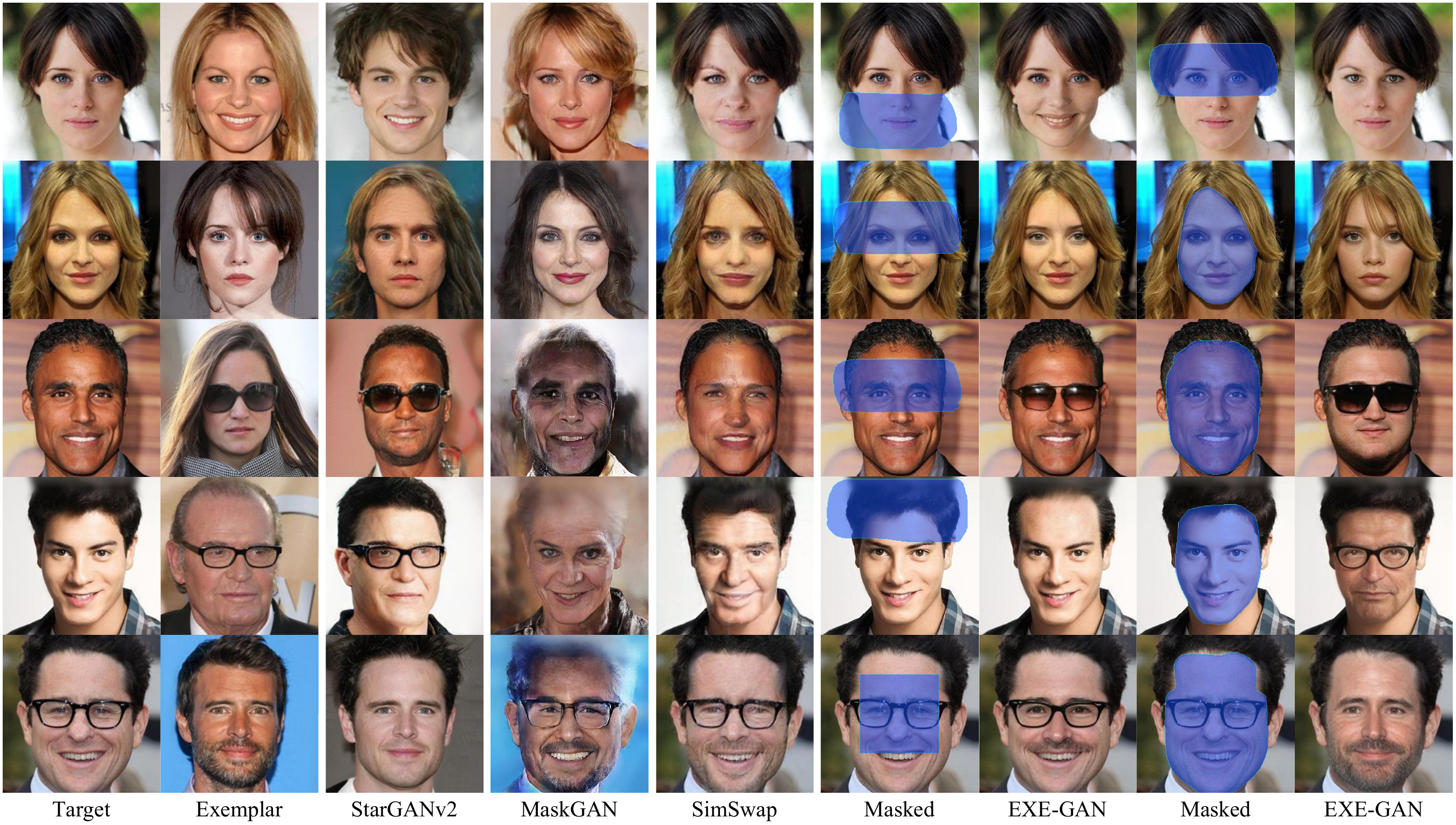}
	\caption{Qualitative comparison between our method and the state-of-the-art facial attribute transfer methods.  }\label{fig:attribute_transfer_comp_sub}  
\end{figure*}

\subsubsection{Quantitative comparison}
Fig.~\ref{fig:facial_inpainting} shows the quantitative comparison of our EXE-GAN to SC-FEGAN~\cite{Jo2019} and LaFIn~\cite{Yang2020}. FID scores with various masked ratios were compared. Experimental results show that our EXE-GAN is able to achieve the best FID scores for all kinds of masks. In terms of the authenticity of inpainted images guided by exemplars, our method outperforms the compared methods.

\subsubsection{Qualitative comparison}
As shown in Fig.~\ref{fig:attribute_inpainting_sub}, SC-FEGAN~\cite{Jo2019} effectively generates facial attributes with shapes guided by sketches but requires more information for high-quality facial attributes inpainting. Moreover, there may be visual artifacts in the inpainted images with SC-FEGAN in Fig.~\ref{fig:attribute_inpainting_sub}. LaFIn~\cite{Yang2020} generates facial expressions similar to exemplars but may fail to inpaint the decorative attributes, such as glasses and hairstyles in Fig.~\ref{fig:attribute_inpainting_sub}. In comparison, our method directly learns the style of facial attributes from the exemplar without extra input, and can successfully generate exemplar-like facial attributes including decorative attributes. As shown in Fig.~\ref{fig:attribute_inpainting_sub}, our EXE-GAN is able to produce more realistic facial inpainted results with facial attributes similar to exemplars.

\subsubsection{User study}\label{sec:guidance_user_study}

We further recruited 63 volunteers to subjectively evaluate the effectiveness of our method comprehensively. 
For the user study, we randomly selected 100 pairs of images from the 550 pairs of images with similar poses mentioned above. Then we randomly divided these selected 100 pairs into 5 groups, and each group consists of 20 pairs with different types of inpainting masks. For each pair, we set one image as the masked input and the other as the exemplar to generate inpainting results using the methods of SC-FEGAN~\cite{Jo2019}, LaFIn~\cite{Yang2020}, and our EXE-GAN, respectively. We then randomly ordered one of these 5 groups to a user. For each round, an exemplar image and its corresponding three inpainting images of SC-FEGAN~\cite{Jo2019}, LaFIn~\cite{Yang2020}, and our EXE-GAN were provided. The participants were asked to select the best image based on the visual quality of inpainting and the perceptual similarity to the exemplar. The results show that our EXE-GAN obtains the majority of votes (59.67\%) compared to SC-FEGAN~\cite{Jo2019} (12.87\%) and LaFIn~\cite{Yang2020} (27.46\%). The user study validates that the output of our EXE-GAN is more realistic than the compared methods visually observed by subjects.

\subsection{Comparison to facial attribute transfer}\label{sec:comparison_transfer}
We compared our method on the Celeba-HQ dataset to the state-of-the-art facial attribute transfer methods, including StarGANv2~\cite{Choi2020}, MaskGAN~\cite{Lee2020}, and SimSwap~\cite{Chen2020_simSwap}, where the same exemplar image was used to guide the attribute transfer.

\begin{figure*}[t]
	\centering
	\includegraphics[width=0.8\textwidth]{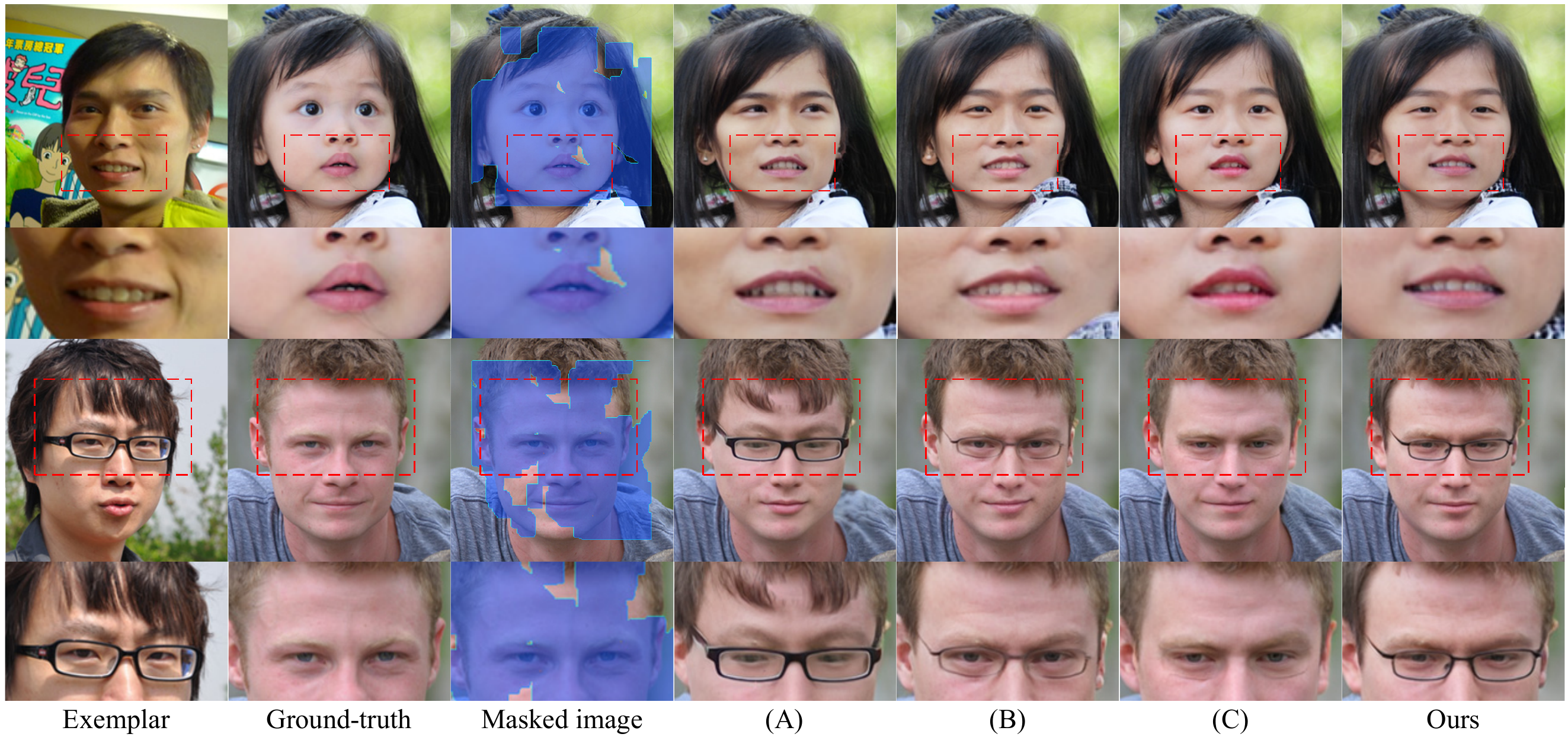}
	\caption{Qualitative examples of the ablation study for large-scale facial inpainting by exemplars with {(A) EXE-GAN without any SVGLs in $\mathcal{L}_{attr}$ and $\mathcal{L}_{lpips}$, (B) EXE-GAN without SVGL in $\mathcal{L}_{attr}$, (C) EXE-GAN  without SVGL in $\mathcal{L}_{lpips}$, and (Ours) EXE-GAN.}}\label{fig:fig_ablation_SVGL}  
\end{figure*}

\subsubsection{Experiment settings}
The pre-trained models of StarGANv2~\cite{Choi2020}, MaskGAN~\cite{Lee2020}, and SimSwap~\cite{Chen2020_simSwap} provided in the official online repository were used in this experiment. For StarGANv2~\cite{Choi2020}, we set the input image as the ``reference'' image and the exemplar image as the ``source'' image. For MaskGAN~\cite{Lee2020}, we extracted semantic masks of input images from the CelebAMask-HQ dataset and obtained the style transferred results based on semantic masks and exemplars. SimSwap~\cite{Chen2020_simSwap} directly performs the exemplar-guided face synthesis with the input and exemplar images. Our EXE-GAN synthesizes facial attributes for masked regions of input images guided by exemplar images.


\begin{table}[t]
	\caption{Ablation study for large-scale facial inpainting with (A) baseline CMOD~\cite{Zhao2021}, (A) CMOD $+$ standard $\mathcal{L}_{lpips}$, and (B) CMOD $+$ SVGL-based $\mathcal{L}_{lpips}$. Results are averaged over 5 runs. \textbf{Bold}: top-2 quantity.}
	\label{tab:ablation_CO_mod_GAN}
	\resizebox{\linewidth}{!}{
		\begin{tabular}{ccccccc}
			\toprule
			\multicolumn{1}{c}{\multirow{2}{*}{Method}} & \multicolumn{3}{c}{CelebA-HQ} &  \multicolumn{3}{c}{FFHQ} \\
			\cline{2-7}    \multicolumn{1}{c}{} & \multicolumn{1}{c}{FID$^{\downarrow}$} & \multicolumn{1}{c}{U-IDS$^{\uparrow}$}& \multicolumn{1}{c}{P-IDS$^{\uparrow}$} & \multicolumn{1}{c}{FID$^{\downarrow}$}
			&\multicolumn{1}{c}{U-IDS$^{\uparrow}$}& \multicolumn{1}{c}{P-IDS$^{\uparrow}$} \\
			\midrule
			CMOD & 9.733  & 9.68\% & 4.00\%  &  4.000  & 26.19\%   &   12.88\%  \\
			(A) &  \textbf{8.993}  &  \textbf{10.87\%}  &  \textbf{4.70\%}  & \textbf{3.432} & \textbf{29.08\%}& \textbf{13.52\%}\\
			(B) & \textbf{8.985}   &\textbf{11.25\%}& \textbf{4.85\%} & \textbf{3.318} &  \textbf{29.97\%} &   \textbf{14.88\%} \\
			\bottomrule
	\end{tabular}}
\end{table}

\subsubsection{Qualitative comparison}
As shown in Fig.~\ref{fig:attribute_transfer_comp_sub}, StarGANv2~\cite{Choi2020} can transform an input image reflecting the identity of the exemplar. However, it leaves users little freedom to manipulate face images interactively. MaskGAN~\cite{Choi2020} transfers the style of exemplar to the input face image using the semantic mask. It requires projecting images into semantic masks and reconstructing images from the mask manifold. As a result, it may introduce irrelevant changes to fine details in the background. SimSwap~\cite{Chen2020_simSwap} can transfer the identity of the exemplar face to the input face and preserve the facial attributes of the input. Nevertheless, it does not allow users to flexibly select regions for face editing. In comparison, our method not only preserves the pixels of known regions but also allows more degrees of freedom to interactively perform the facial attribute manipulation. Our EXE-GAN is able to produce 
high-quality results with facial attributes guided by exemplars, including gender, makeup style, hairstyle, and decorative style (e.g., glasses).

\begin{figure*}[t]
	\centering
	\includegraphics[width=\textwidth]{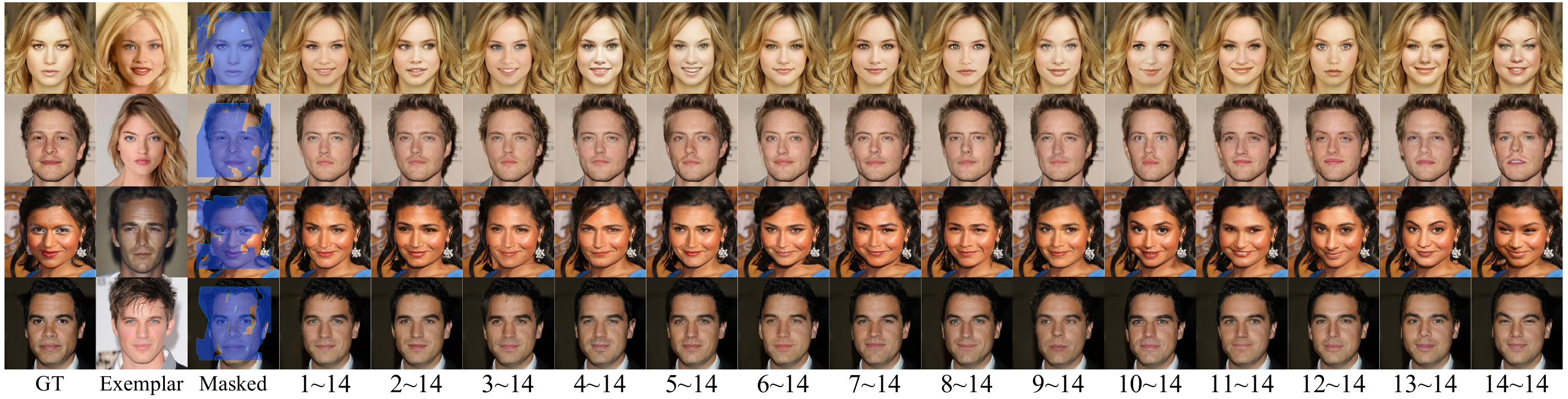}
	\caption{{Examples of facial inpainting with various subsets of the style codes: ground-truth, exemplar, masked image, and various style effects. In each row, values of $i$-th to $j$-th layers in the style code are from the exemplar and values of remaining layers are from the stochastic style code.}}\label{fig:fig_style_code_ablation}  
\end{figure*}

\subsection{Ablation study}

\begin{table}[t] 
	\caption{Ablation study for large-scale facial inpainting by exemplars with {(A) EXE-GAN without any SVGLs in $\mathcal{L}_{attr}$ and $\mathcal{L}_{lpips}$, (B) EXE-GAN without SVGL in $\mathcal{L}_{attr}$, (C) EXE-GAN  without SVGL in $\mathcal{L}_{lpips}$, and (Ours) EXE-GAN.} Results are averaged over 5 runs. \textbf{Bold}: top-2 quantity.}
	\label{tab:ablation_quantitative}
	\resizebox{\linewidth}{!}{
		\begin{tabular}{ccccccc}
			\toprule
			\multicolumn{1}{c}{\multirow{2}{*}{Method}} & \multicolumn{3}{c}{CelebA-HQ} &  \multicolumn{3}{c}{FFHQ} \\
			\cline{2-7}    \multicolumn{1}{c}{} &  \multicolumn{1}{c}{FID$^{\downarrow}$} & \multicolumn{1}{c}{U-IDS$^{\uparrow}$}& \multicolumn{1}{c}{P-IDS$^{\uparrow}$} 
			& \multicolumn{1}{c}{FID$^{\downarrow}$}
			&\multicolumn{1}{c}{U-IDS$^{\uparrow}$}& \multicolumn{1}{c}{P-IDS$^{\uparrow}$} \\
			\midrule
			(A) &  12.853   & 4.024\% &  1.25\%	& {7.298}  & {17.29\%} & {6.55\%}\\
			(B)  & 10.433  &  \textbf{8.875\%} & 3.35\% & 4.909  & 22.46\% & 8.75\% \\
			(C) &  \textbf{9.804}   & \textbf{8.875\%} &  \textbf{4.25\%}	& \textbf{4.408}  & \textbf{24.61\%} & \textbf{10.04\%}\\
			Ours &  \textbf{9.967} &  \textbf{9.175\%}  & \textbf{3.85\%}  & \textbf{4.353}  & \textbf{24.33\%} &  \textbf{9.92\%} \\
			\bottomrule
	\end{tabular}}
\end{table}

\subsubsection{{Ablation study on SVGL in CMOD~\cite{Zhao2021}}}
To demonstrate the effectiveness of the proposed spatial variant gradient layer (SVGL), we experimented to apply our SVGL to CMOD for the image inpainting task. As shown in Table~\ref{tab:ablation_CO_mod_GAN}, the performance is improved by adding the LPIPS loss into the baseline CMOD. The quantitative scores are considerably improved by further introducing SVGL in the LPIPS loss. The SVGL-based LPIPS loss helps the generator focus more on pixels close to the hole boundary to avoid visual inconsistency and still encourage inherent stochasticity with less constraints to pixels away from the boundary. This ablation study shows that the proposed SVGL can effectively boost the performance of image inpainting.

\subsubsection{{Ablation study on SVGL in EXE-GAN}}
We further investigated the effectiveness of each component in our EXE-GAN by performing ablation study. 
Table~\ref{tab:ablation_quantitative} shows the quantitative results. We also present qualitative exemples in Fig.~\ref{fig:fig_ablation_SVGL} to better express visual effects of the ablation study. {When removing SVGLs in both the attribute loss and the LPIPS loss (A), both the quantitative measures and visual qualities drop dramatically.} When replacing the SVGL-based attribute loss with the standard attribute loss without SVGL (B), there may be visible boundary inconsistencies in the generated results, and the quantitative performance is also affected. When replacing the SVGL-based LPIPS loss with the standard LPIPS loss without SVGL (C), the visual similarities of facial attributes (e.g., facial expression, wearing glasses) between the generated result and the exemplar image decrease, while the quantitative scores are comparable to EXE-GAN for both testing datasets. In this case, the standard LPIPS loss is applied to all pixels of the image to enforce the generator to reconstruct the contents of ground-truth instead of exemplar attributes, as demonstrated in Fig.~\ref{fig:fig_ablation_SVGL}. In comparison, our EXE-GAN is able to produce realistic facial images with facial attributes similar to exemplars and competitive quantitative scores.

\begin{figure}[t]
	\centering
	\includegraphics[width=0.5\textwidth]{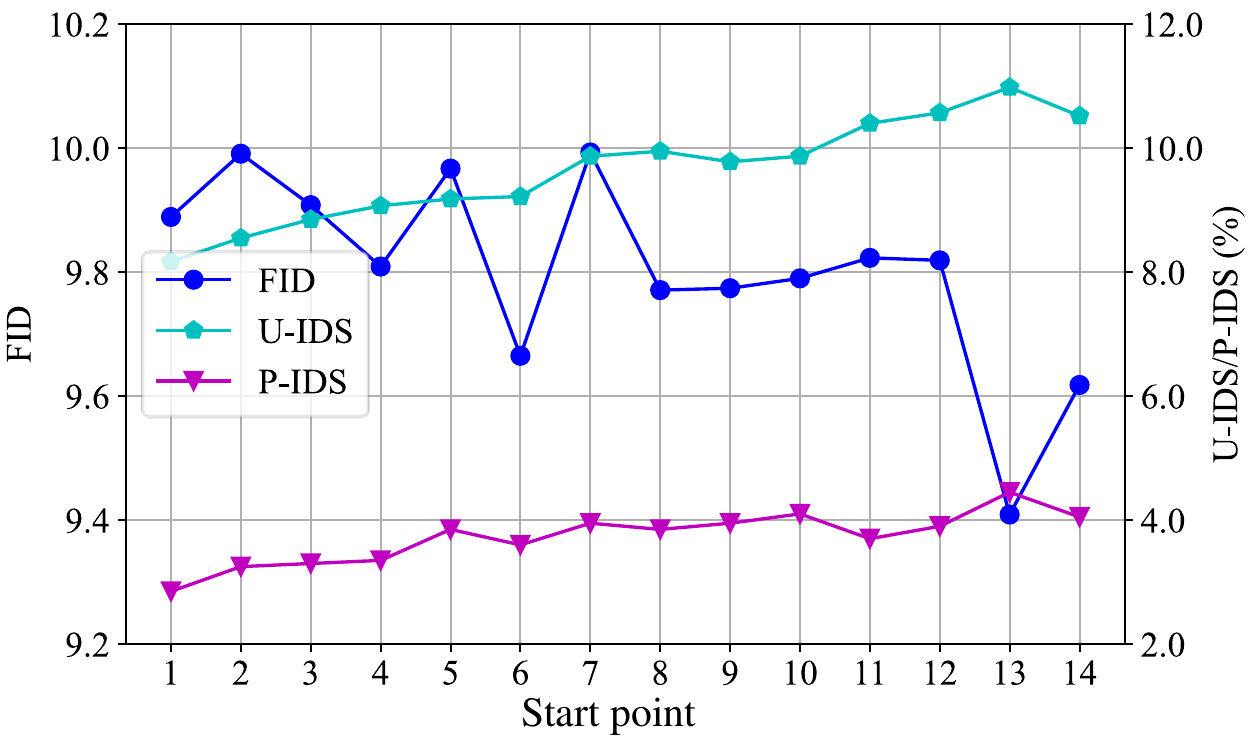}
	\caption{{Ablation study for large-scale facial inpainting by exemplars on modulated exemplar style codes on the CelebA-HQ dataset. The start point $i$ indicates $i$-{th} to 14-{th} style
			layers are modulated with exemplar styles while the other layers are with random styles.}}\label{fig:fig_style_code_ablation_quantitative}  
\end{figure}

\begin{figure*}[t]
	\centering
	\includegraphics[width=\textwidth]{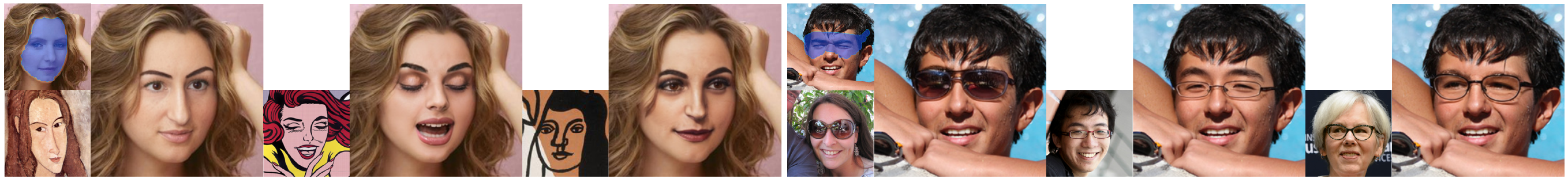}
	\caption{Examples of local facial attribute transfer guided by different cartoon exemplar images (left group) and real-world exemplar images (right group).}\label{fig:fig_local_edit}  
\end{figure*}

\begin{figure*}[t]
	\centering
	\includegraphics[width=\textwidth]{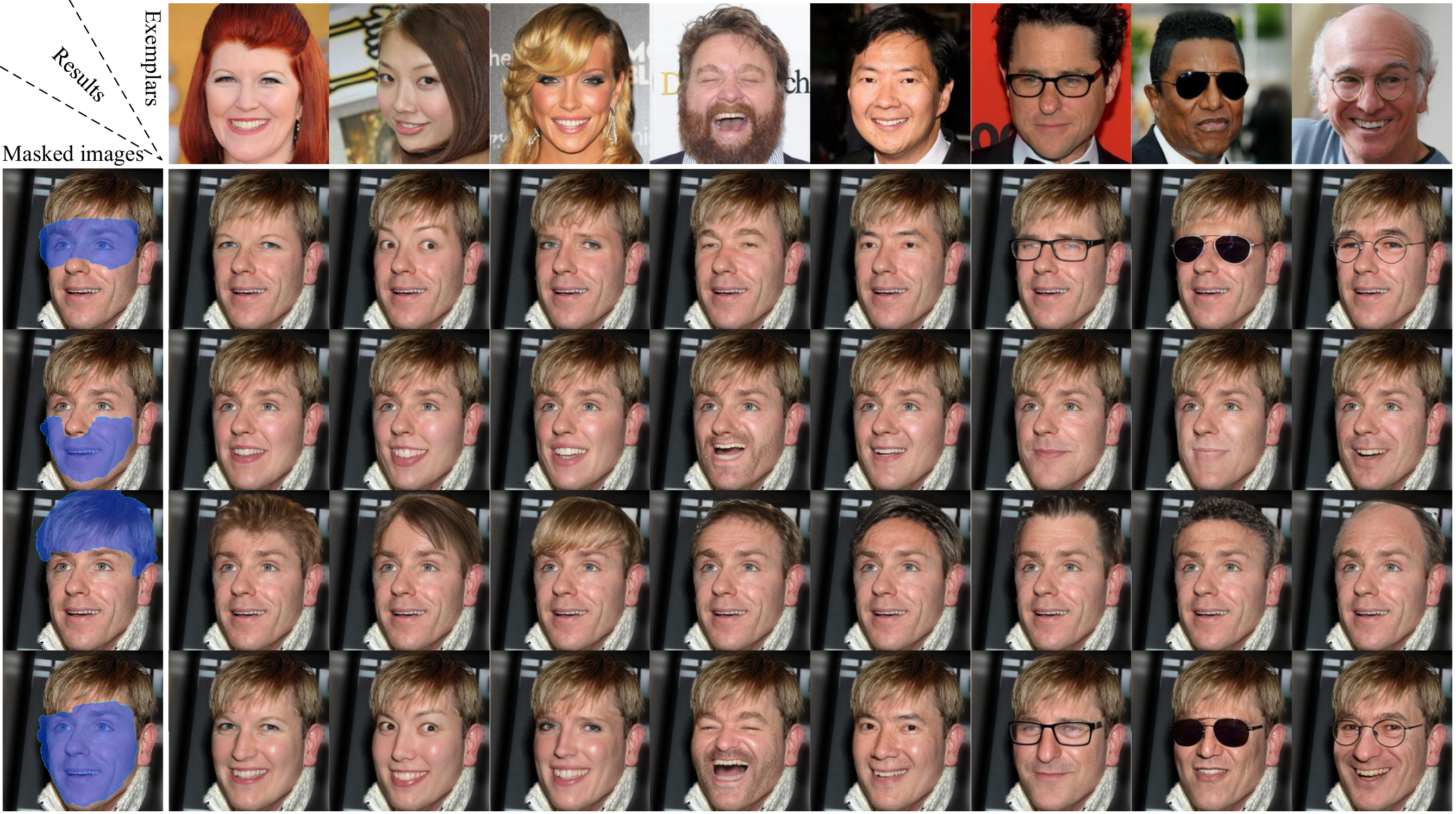}
	\caption{More examples of local facial attribute transfer guided by exemplars.}\label{fig:fig_local_edit_diverse}  
\end{figure*}

\subsubsection{{Ablation study on exemplar styles modulation}}
{We conducted ablation experiments on the exemplar styles modulation by re-training our model with various configurations of the vector $\phi$. As shown in Eq.~\ref{equ:mixing}, $\phi$ is a binary vector to indicate which style is modulated for each style layer. From Fig.~\ref{fig:fig_style_code_ablation} we can find that the more exemplar style codes are modulated the more exemplar facial attributes will present in the inpainted images. We recorded the quantitative scores for various subsets of the style codes in Fig~\ref{fig:fig_style_code_ablation_quantitative}. The quantitative scores gradually get better with slight fluctuations when fewer exemplar styles are modulated. This ablation study validated that more exemplar styles lead to more exemplar facial attributes in inpainted images while the quantitative performance decreases.}


\begin{figure*}[t]
	\centering
	\includegraphics[width=\textwidth]{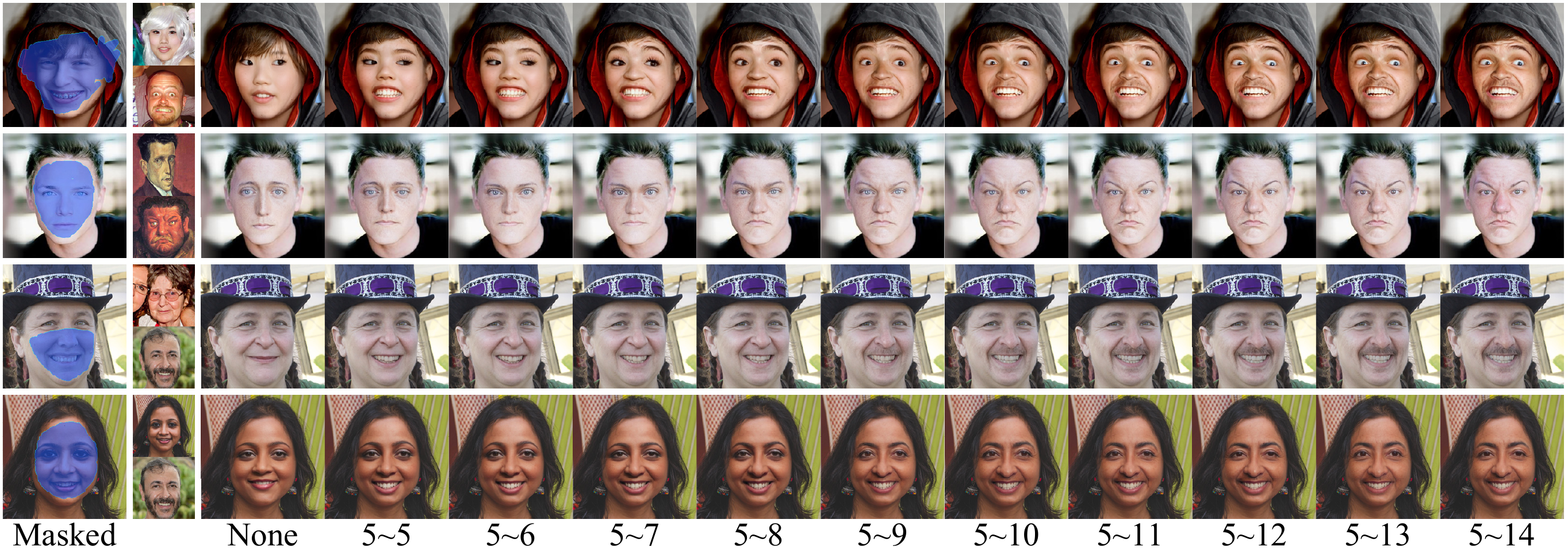}
	\caption{Examples of guided facial style mixing (from left to right in each group): masked image, pairs of exemplars, and style-mixing effects. In each row, values of the style code of the first exemplar from $i$-th to $j$-th layers are replaced by those of the second exemplar.}\label{fig:fig_style_mixing_diverse_sub}  

	\centering
	\includegraphics[width=\textwidth]{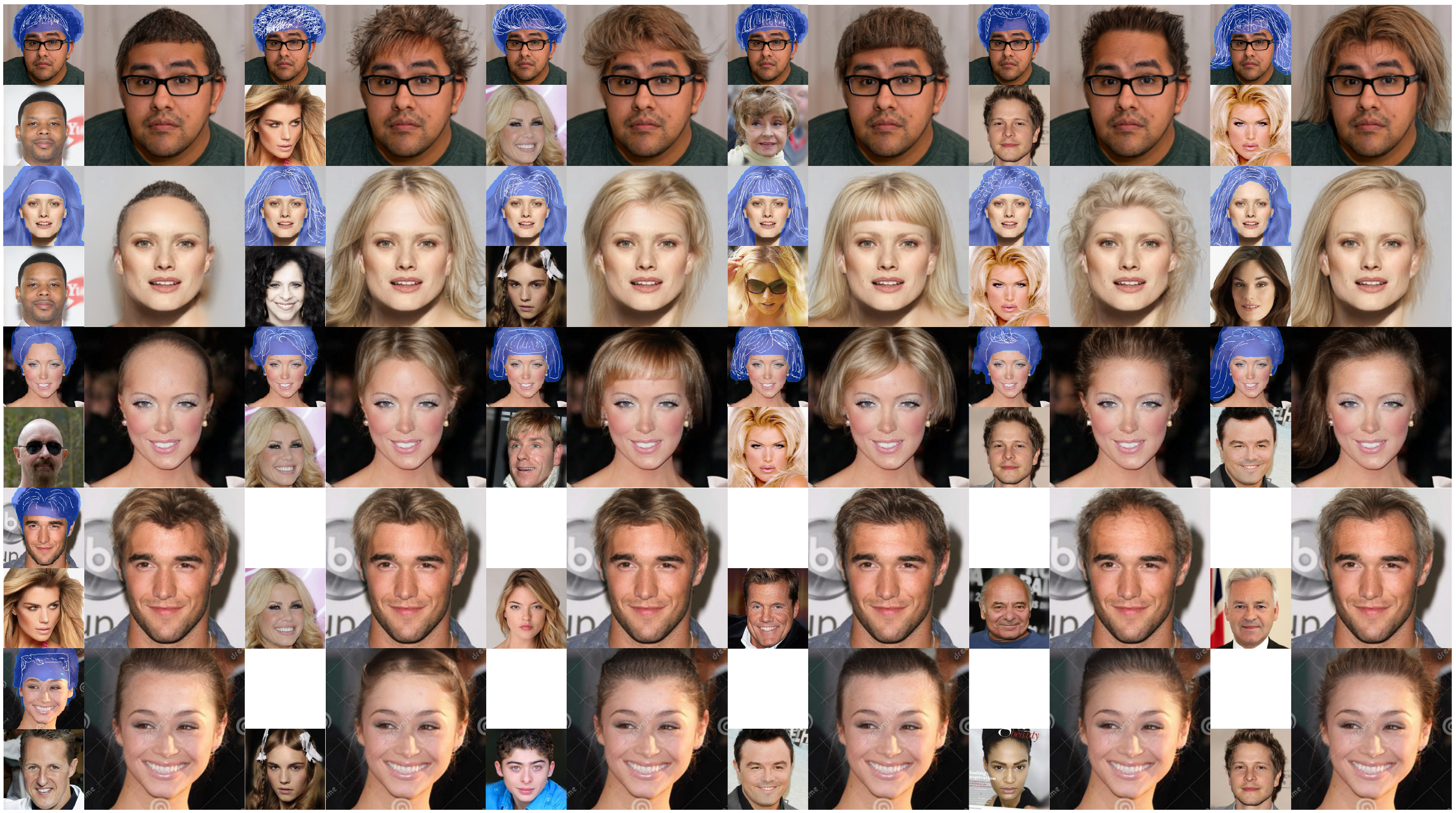}
	\caption{Examples of hairstyle editing: (top three rows) edited results with different sketches guided by different exemplars and (last two rows) edited results with the same sketch guided by different exemplars.}\label{fig:fig_hair_style_edit}  
\end{figure*}

\section{Applications with our facial inpainting}
In this section, we demonstrate a number of applications by equipping our generative facial inpainting method.


\subsection{Local facial  attribute transfer}
Since our EXE-GAN helps the generator learn the mapping between injected exemplar representations and corresponding facial attributes, our method can be used to produce vivid facial attribute transfer effects guided by various exemplars, such as real-world facial attributes and artistic expressions. As shown in Fig.~\ref{fig:fig_local_edit} and Fig.~\ref{fig:fig_local_edit_diverse}, for a masked input, our EXE-GAN produces high-quality local facial attribute transfer results by leveraging facial attributes of exemplars. 

\begin{figure*}[t]
	\centering
	\includegraphics[width=\textwidth]{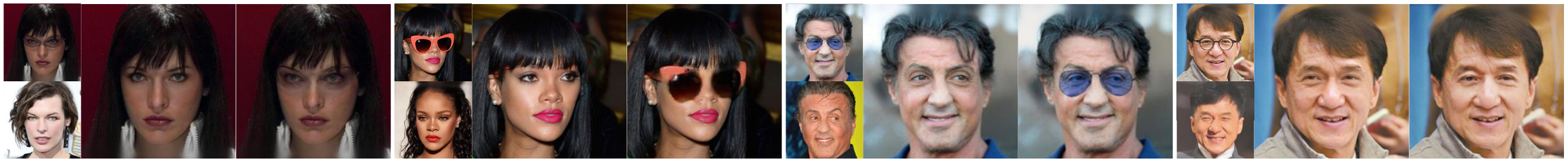}
	\caption{Comparison on portrait eyeglasses removal (from top left
		to right in each group): masked image, exemplar image, our recovered result, and Lyu et al.'s result~\cite{Lyu2022}. }\label{fig:fig_eyeglass_remove}  
\end{figure*}

\subsection{Guided facial style mixing}
This application shows that our EXE-GAN can be used to generate facial inpainting effects by mixing two exemplar style latent codes. We first employ the style encoder $E$ to obtain two exemplar style codes from two exemplars, respectively. Then, we apply the style mixing~\cite{Karras2019,Karras2020} on the two latent codes with a crossover point. By simply changing the crossover point, we can obtain multiple mixed latent codes.  Therefore, guided facial style mixing effects can be obtained by moving the crossover point over the vector $\phi$ in Eq.~\ref{equ:mixing}, as shown in Fig.~\ref{fig:fig_style_mixing_diverse_sub}.

\subsection{Hairstyle editing}
In this application, we further fine-tuned our trained EXE-GAN with extra hand-drawn-like sketches which were produced automatically with a pencil-sketch filter~\cite{Richardson2021}. The application allows users to sketch in the mask to indicate roughly the hair styles. Given a masked image with sketches and an exemplar image, our EXE-GAN produces a style edited output. As shown in Fig.~\ref{fig:fig_hair_style_edit}, various hairstyle editing results are obtained by changing the user-edited style sketches. It is easy even for a novice to obtain various styles by simple sketch editing.

\begin{figure*}[t]
	\centering
	\includegraphics[width=\textwidth]{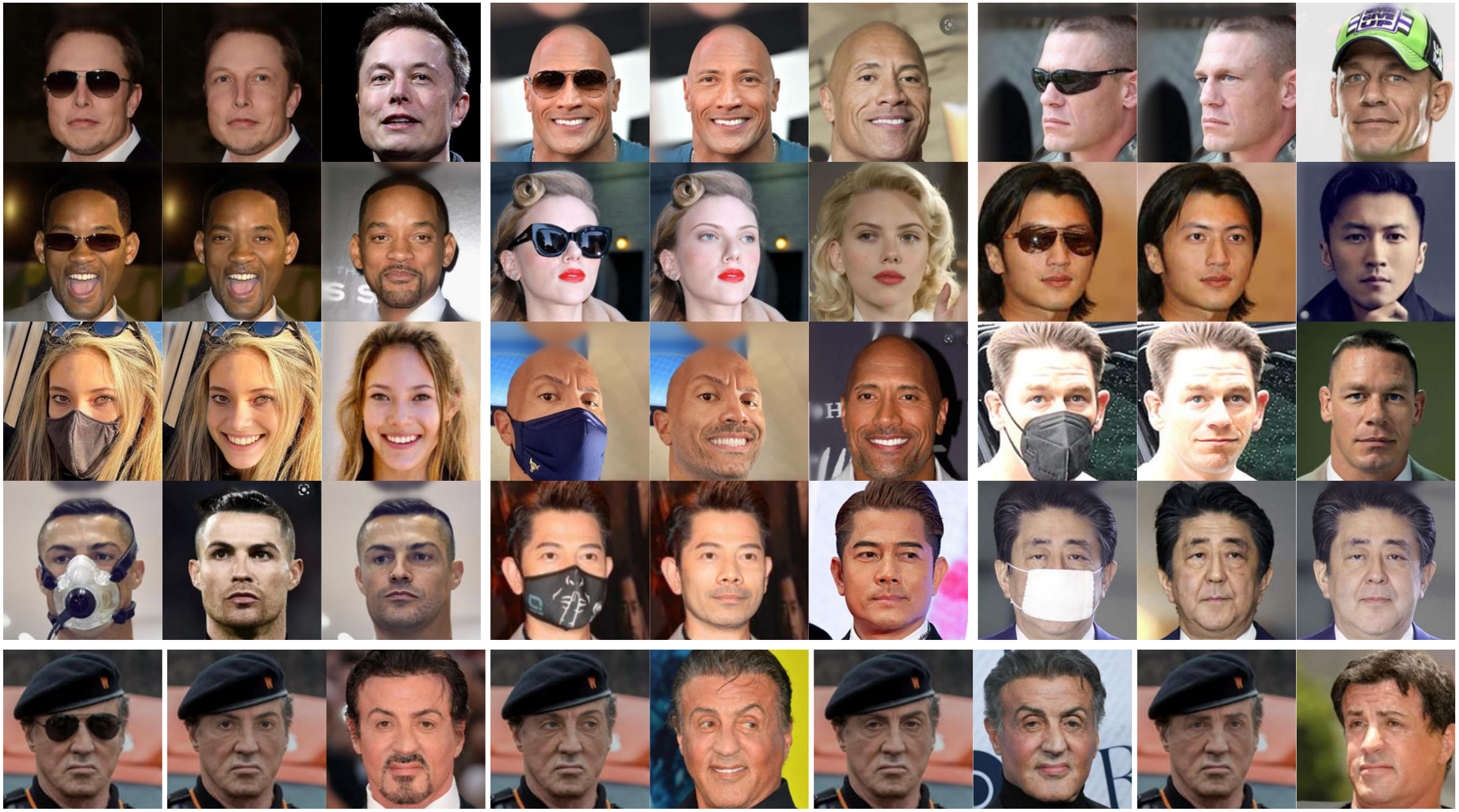}
	\caption{More examples of guided facial image recovery: (top four rows, from left to right in each group) occluded image, recovered face image, and exemplar; (bottom row) diverse recovered results guided by different exemplars.}\label{fig:fig_facial_reconstruction}  
\end{figure*}

\begin{figure*}[t]
\centering
\includegraphics[width=\textwidth]{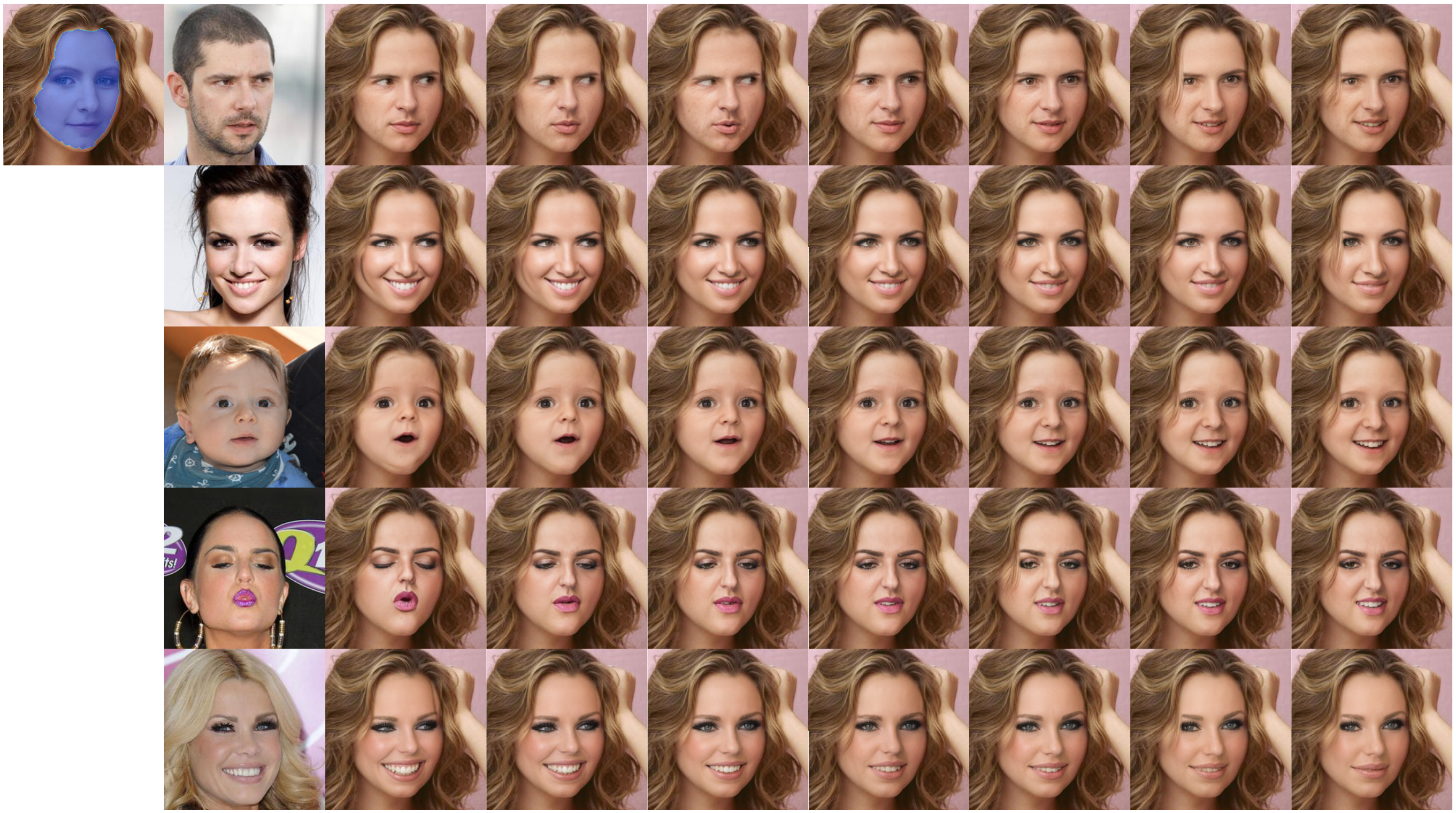}
\caption{Examples of diverse facial inpainting with inherent stochasticity. Based on the same masked image, we use different exemplars to guide the generation of various results. }\label{fig:fig_inherent_diverse_appendix}  
\end{figure*}

\subsection{Guided facial image recovery}
The guided facial image recovery for occluded portrait eyeglasses and masks was implemented by masking them out and taking a different image from the same person as exemplar. As shown in Fig.~\ref{fig:fig_eyeglass_remove}, we compared our method to Lyu et al.'s eyeglasses removal method~\cite{Lyu2022} on tinted eyeglasses (leftmost), sunglasses (mid-left), and myopia glasses (mid-right). The results show that our guided facial image recovery method performs well on removing tinted eyeglasses and sunglasses which may fail with Lyu et al.'s method. As show in Fig.~\ref{fig:fig_facial_reconstruction}, our method can also recover faces from occluded masks effectively.


\subsection{Inherent stochasticity}
Our method can be extended to produce multiple diverse facial inpainting results for an input masked facial image and an exemplar image by leveraging the inherent stochasticity. Users can easily select the preferred one among these results. The inherent stochasticity is achieved by adding per-pixel noise after each convolutional layer, {leveraging the injected random latent code}, and applying a truncation trick to tune the stochastic style representations~\cite{Karras2019,Karras2020}. Fig.~\ref{fig:fig_inherent_diverse_appendix} shows a variety of facial inpainting results {with various random latent codes.}


%

\section{Conclusion, limitations, and future work}\label{sec:con}
In this paper, we have presented a novel interactive framework for realistic facial inpainting by taking advantages of exemplar facial attributes. An attribute similarity metric was introduced to help the generative network learn the style of facial attributes from the exemplar. We further proposed a novel spatial variant gradient backpropagation technique to address the issue of visual inconsistency on the filling boundary. Extensive experiments and applications have demonstrated the effectiveness of the proposed method.

{Our method has some limitations. Using the embedded style codes, we successfully transfer the facial attribute styles from the exemplar image. The explicit mapping between the facial attribute and the embedded style codes, on the other hand, is still unknown ~\cite{Richardson2021}. Incorporating a more advanced embedding algorithm into our pipeline could be a good next step.  The trained model works well for aligned images because the facial images in the experimented training datasets~\cite{Karras2018,Karras2019} are highly aligned. It is necessary to align and crop the inputs before inpainting nonaligned images. It would be preferable to train models on unstructured datasets to create a more sophisticated algorithm.}
\ifCLASSOPTIONcaptionsoff
  \newpage
\fi



\bibliographystyle{IEEEtran}
\bibliography{IEEEabrv,guided_inpainting_bib}
\end{document}